\documentclass[10pt,twocolumn,letterpaper]{article}

\usepackage{cvpr}              
\usepackage{algorithm}
\usepackage[utf8]{inputenc}
\usepackage{algpseudocode} 
\usepackage{amsmath}
\usepackage{amssymb}
\usepackage{booktabs} 
\usepackage{listings} 

\usepackage{listings} 
\usepackage{mdframed} 
\usepackage[T1]{fontenc} 
\usepackage{textcomp}     
\definecolor{cvprblue}{rgb}{0.21,0.49,0.74}
\usepackage[pagebackref,breaklinks,colorlinks,allcolors=cvprblue]{hyperref}

\def\paperID{43902} 
\def\confName{CVPR}
\def\confYear{2026}

\definecolor{r4color}{RGB}{0, 0, 0}

\title{HOG-Layout: Hierarchical 3D Scene Generation, Optimization and Editing via Vision-Language Models}

\author{
Haiyan Jiang$^{1}$ \hspace{0.8em} Deyu Zhang$^{1,2}$ \hspace{0.8em} Dongdong Weng$^{2}$ \hspace{0.8em} Weitao Song$^{2}$ \footnotemark[1] \hspace{0.8em} Henry Been-Lirn Duh$^{1}$\\
$^{1}$The Hong Kong Polytechnic University \hspace{0.8em}
$^{2}$Beijing Institute of Technology\\
}

\begin{document}

\twocolumn[{%
\renewcommand\twocolumn[1][]{#1}%
\maketitle
\begin{center}
\centering
\vspace{-0.5cm}
\includegraphics[width=0.85\textwidth]{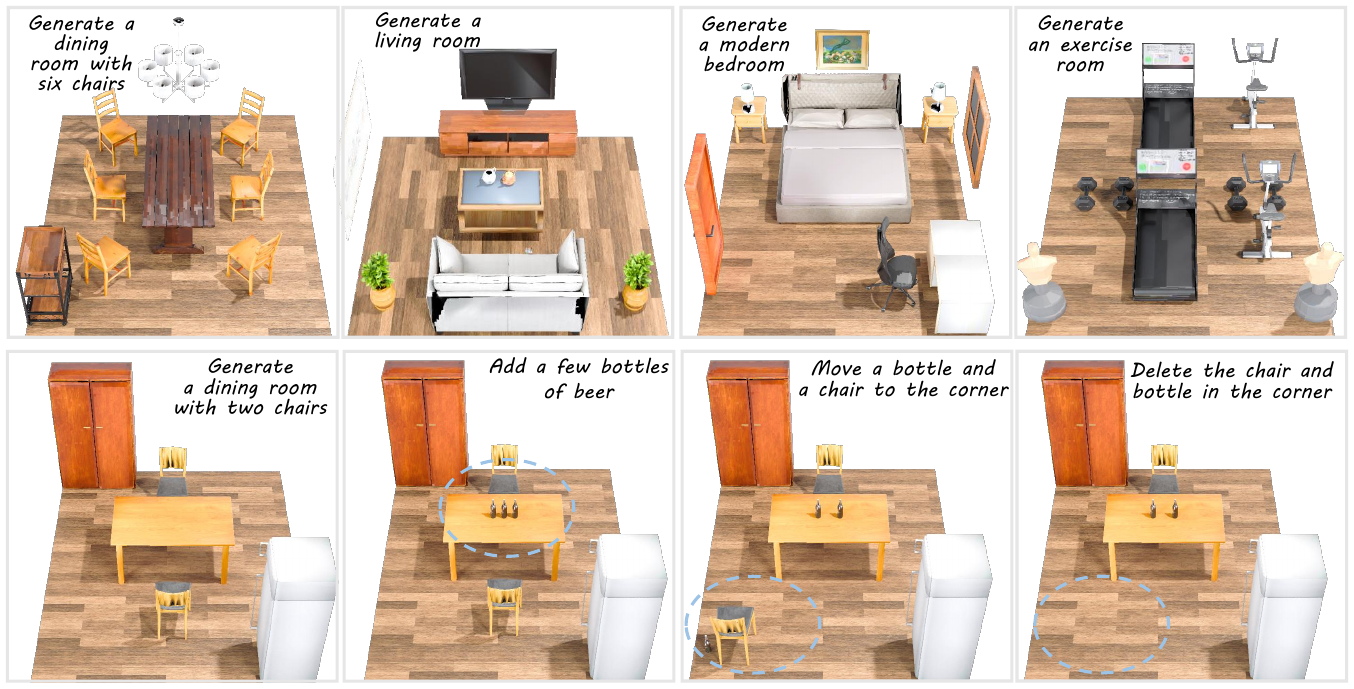}
\vspace{-0.2cm}
\captionof{figure}{Scene generation and editing examples from HOG-Layout, which can provide physically plausible and semantically coherent scenes through hierarchical generation and optimization.}
\end{center}%
}]
\begingroup
\renewcommand{\thefootnote}{\fnsymbol{footnote}}
\footnotetext[1]{~Corresponding author. \textit{Email: swt@bit.edu.cn.}}
\endgroup

\lstset{
  basicstyle=\scriptsize\ttfamily,
  breaklines=true,
  tabsize=4,
  showstringspaces=false,
  showspaces=false,
  showtabs=false,
  frame=none,
  backgroundcolor=\color{gray!10},
}

\newcommand{\styledfileinput}[3][]{%
  \begin{mdframed}[backgroundcolor=gray!10,linewidth=0pt]
    \lstinputlisting[language=#1,escapeinside={(*@}{@*)},basicstyle=\scriptsize\ttfamily,#3]{#2}
  \end{mdframed}
}

\renewcommand{\algorithmicrequire}{\textbf{Input:}}
\renewcommand{\algorithmicensure}{\textbf{Output:}}

\begin{abstract}
3D layout generation and editing play a crucial role in Embodied AI and immersive VR interaction. However, manual creation requires tedious labor, while data-driven generation often lacks diversity. The emergence of large models introduces new possibilities for 3D scene synthesis. We present HOG-Layout that enables text-driven \underline{\textbf{h}}ierarchical scene \underline{\textbf{g}}eneration, \underline{\textbf{o}}ptimization and real-time scene editing with large language models (LLMs) and vision-language models (VLMs). HOG-Layout improves scene semantic consistency and plausibility through retrieval-augmented generation (RAG) technology, incorporates an optimization module to enhance physical consistency, and adopts a hierarchical representation to enhance inference and optimization, achieving real-time editing. Experimental results demonstrate that HOG-Layout produces more reasonable environments compared with existing baselines, while supporting fast and intuitive scene editing.
\end{abstract} 
\section{Introduction}
\label{sec:intro}

3D indoor scene generation enables rapid synthesis of realistic, interactive, and diverse environments, serving as a foundation for interior design, 3D games, virtual reality, and embodied AI. High-quality scene generation relies on both reasonable layout planning and appropriate object selection. Traditional methods typically learn scene layouts from data~\cite{zhou2019scenegraphnet,wang2019planit,li2019grains,luo2020end,purkait2020sg,ritchie2019fast,paschalidou2021atiss,para2023cofs,para2021generative}, or directly generate layout appearances~\cite{fang2025ctrl, epstein2024disentangled,po2024compositional,schult2024controlroom3d,zhou2024dreamscene360,Zhou2024GALA3DTT}, which are limited in diversity or lack interactivity.

The emergence of LLMs and VLMs has facilitated more flexible scene generation. LLMs can infer object sizes, positions, and orientations from text~\cite{feng2023layoutgpt}, but may produce collisions or unreasonable placements. Spatial relation constraints have been introduced to improve scene plausibility~\cite{yang2024holodeck}, though at the cost of generative diversity. VLMs leverage both textual and visual information to generate layouts that respect spatial distribution~\cite{deng2025global,sun2025layoutvlm}, and methods like LayoutVLM optimize object placement within predefined asset groups to enhance physical plausibility, yet still require prior specification of the object set. Moreover, existing approaches primarily focus on generating scenes from scratch, neglecting scene editing, which is crucial in practice. Users often wish to modify existing scenes by moving, adding, or removing objects rather than rebuilding entirely.

Therefore, we propose HOG-Layout, a hierarchical scene generation and optimization framework that also supports editing. Objects are organized hierarchically according to support relationships, e.g., all objects directly contacting the floor form the same level, with the floor as the parent and supported objects as children. During the optimization, relationships within the same level and parent-child connections are considered and iteratively optimized, with optimization occurring at multiple levels simultaneously. Forces are decomposed into vertical forces, horizontal forces, and rotations to guide the object's movement, thereby improving optimization efficiency. A template rule library combined with retrieval-augmented generation (RAG) technology further enhances layout plausibility. Specifically, user text is first vectorized and combined with templates via RAG to generate layout planning and detailed scene descriptions. These, along with top-down views of the scene~\cite{yang2023set,cai2024vip,lei2025scaffolding}, are input to a VLM to produce an initial layout, followed by hierarchical physical and semantic optimization. For scene editing, LLM parses the text into specific operations (move, add, delete) and combines it with existing scene information to generate an initial layout through VLM, which is then optimized to obtain the final edited scene. Overall, our contributions are as follows: 

1) Propose a hierarchy-based multi-step scene layout generation, optimization, and editing method that enhances VLM reasoning capabilities to generate interactive scenes. 

2) Introduce a hierarchical optimization module that decomposes force directions to achieve optimization between same-level objects and parent-child related objects, enabling fast and precise optimization. 

3) Experiments demonstrate that HOG-Layout can rapidly generate more reasonable scene layouts and achieve low-latency scene editing.
\section{Related work}
\label{sec:relatedwork}

\subsection{3D Indoor Scene Synthesis}

Indoor scenes can be obtained through 3D scans of real-world environments~\cite{ramakrishnan2021habitat, hua2016scenenn}, NeRF or Gaussian splat rendering~\cite{epstein2024disentangled, po2024compositional, schult2024controlroom3d, zhou2024dreamscene360, Zhou2024GALA3DTT}, or generated via GANs~\cite{bahmani2023cc3d}. However, objects in these scenes are difficult to segment or manipulate, limiting their use in Embodied AI or virtual reality. Data-driven methods generate layouts via graphs~\cite{zhou2019scenegraphnet, wang2019planit, li2019grains, luo2020end}, procedural grammars~\cite{purkait2020sg}, auto-regressive networks~\cite{ritchie2019fast}, Transformers~\cite{paschalidou2021atiss,para2023cofs,para2021generative}, or diffusion models~\cite{fang2025ctrl}, often retrieving objects from asset libraries. While these approaches support interactive scenes, they remain heavily dependent on training data, limiting generative diversity. Recently, large language and multimodal models have leveraged their powerful reasoning capabilities to enable dataset-free scene generation, which will be detailed in the following section.

\subsection{Text-driven 3D Scene Generation}

Language, as an important medium for human–computer interaction, is widely used in 3D scene generation. In recent years, researchers have leveraged diffusion models, semantic graph priors, and multimodal generative frameworks to achieve automatic construction and control of 3D scenes from text descriptions~\cite{tang2024diffuscene,bautista2022gaudi,lin2024instructscene,song2023roomdreamer,fang2025ctrl}. These methods can generate scenes consistent with textual semantics in terms of geometry, semantics, and layout, and also support scene editing and manipulation through linguistic instructions~\cite{chang2017sceneseer,ma2018language,chang2015text}. However, such methods still face limitations in semantic consistency, generative diversity, and generalization capability. Compared with these methods, large models enable more flexible 3D scene generation with open-vocabulary without relying on datasets. Recent studies show that LLMs can directly produce reasonable scene layouts~\cite{feng2023layoutgpt,littlefair2025flairgpt}, or convert textual narratives into generative scene representations through structured forms~\cite{yang2024holodeck,ccelen2024design}. Moreover, VLMs can jointly leverage textual and visual information to generate layouts that better align with semantic intent and spatial constraints~\cite{deng2025global,sun2025layoutvlm}. However, current methods still face limitations in physical consistency, interaction controllability, and computational efficiency.

\subsection{3D Reasoning with Vision-Language Models}

Vision Language Models (VLMs) possess spatial reasoning capabilities, and prior research has leveraged them for tasks such as scene understanding, question answering, navigation, and planning~\cite{hong20233d,yang20253d,chen2024ll3da,qi2024gpt4point,huang2024chat}. While standard input-output paradigms~\cite{Qwen-VL,liu2023visual} perform well on simple tasks like question answering, their effectiveness becomes limited when tasks grow more complex. Consequently, researchers have adopted methods such as Chain-of-Thought (CoT) and Tree-of-Thought (ToT)~\cite{wei2022chain,yao2023tree} to enable step-by-step reasoning and tackle complex tasks. Furthermore, visual markers~\cite{yang2023set,cai2024vip,lei2025scaffolding,sun2025layoutvlm} can reduce VLMs' reliance on textual cues, facilitate interpretation of spatial layouts, and enhance their ability to reason about spatial relationships.

In this work, we leverage the spatial reasoning capabilities of VLM, combining the top-down layout view with visual markers to generate more reasonable 3D layouts while enabling low-latency real-time editing.
\section{HOG-Layout}
\label{sec::methods}

\begin{figure*}[t]
  \centering
   \includegraphics[width=1\linewidth]{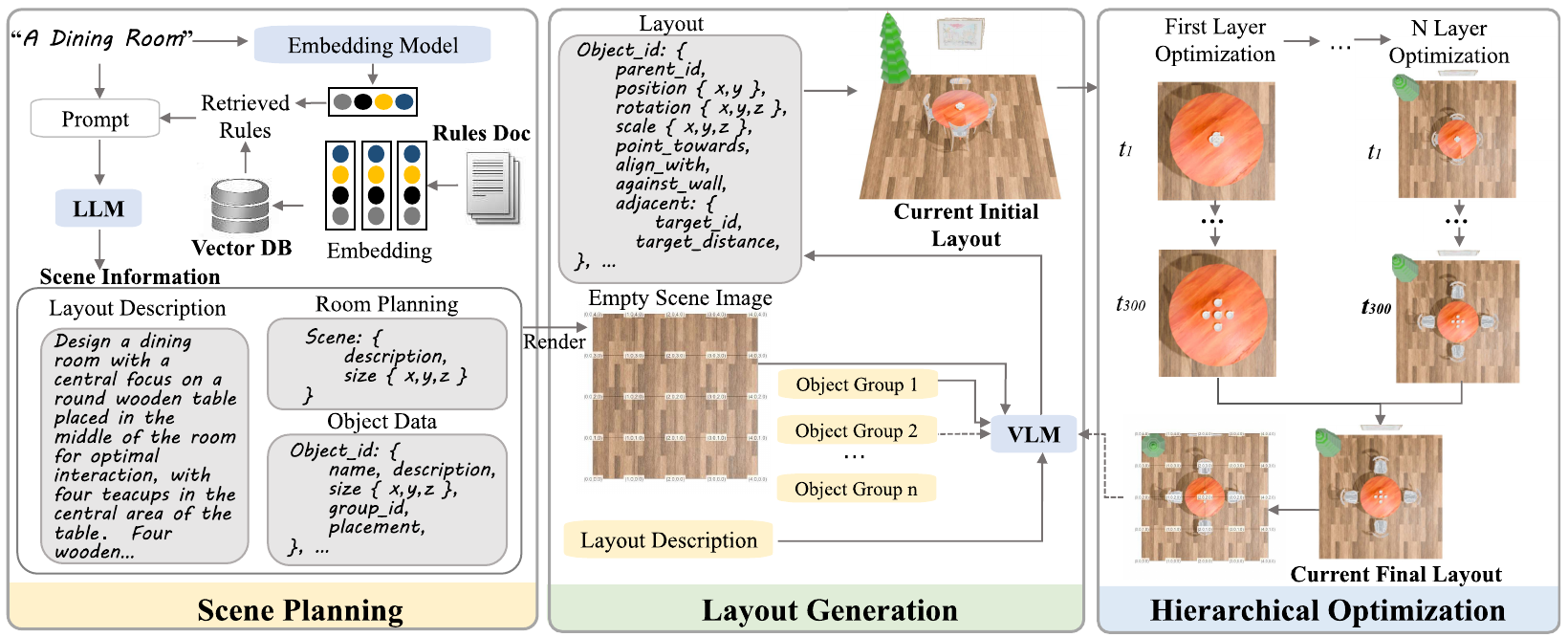}
   \caption{The pipeline of HOG-Layout. In the layout generation phase, layouts are generated sequentially according to groups and then optimized. The optimized scene is used as input for generating the layout for the next group. In the optimization phase, iterative optimization is performed according to the parent-child hierarchy, with optimization occurring simultaneously across different layers.}
   \label{fig:pipeline}
\end{figure*}

HOG-Layout is a modular pipeline for scene generation and editing, consisting of four key components: (1) \textbf{Scene Planning}, where an LLM with RAG produces a structured plan including object lists, room information, and overall scene description; (2) \textbf{Layout Generation}, where a VLM integrates the scene plan with a top-down view annotated with visual markers to generate a hierarchical layout and instantiate objects to generate the initial scene; (3) \textbf{Hierarchical Optimization}, which iteratively refines object positions and orientations under structured constraints to obtain the final scene; and (4) \textbf{Scene Editing}, supporting precise text-guided modifications. More details could be found in the supplementary materials.

\subsection{Scene Planning}

As shown in \Cref{fig:pipeline}, the Scene Planning module aims to identify the objects required for a target scene from a natural language description, generate corresponding layout guidance, and provide an overall scene description, outputting the scene planning in textual form.

Within this module, we construct a template library of layout constraint rules to improve the plausibility of generated layouts. The templates are first segmented into fixed-size text blocks. For each block, we extract a 1024-dimensional feature vector using a text embedding model (in this work, Qwen3-Embedding-4B\cite{zhang2025qwen3} is used) and store the vectors in a FAISS vector database~\cite{johnson2019billion}. \textcolor{r4color}{More details could be found in the supplementary materials.} 

When a user provides a text input, the input is first embedded using the same text embedding model, and its cosine similarity with the vectors in the database is computed to retrieve the three most relevant layout rules. These retrieved rules, along with the input text, are then fed into the LLM prompt to directly generate the layout plan, including object-level information, the target scene context description and scene size information. Each object entry contains its ID, name, description, size, placement and group. The group refers to different functional areas. For example, a combined bedroom and living room space can be divided into two groups: a dining group and a movie-watching group. Scene size information includes the length, width, and height of the room, which can be used for rendering the top view of the scene later.

\subsection{Layout Generation}

As shown in \Cref{fig:pipeline}, the \textbf{Layout Generation} module embeds the scene plan into the VLM prompt and includes a top-down view of the current scene with grid lines and coordinates. Layouts are generated iteratively for multiple groups: the first group uses an empty-scene view, while subsequent groups use the top-down view of the currently optimized scene. After generating the textual layout for each group, object retrieval is performed, followed by refinement of object positions and orientations by the \textbf{Hierarchical Optimization} module to produce the optimized scene. This optimized scene then serves as input for generating the next group’s layout, enabling sequential and consistent scene construction until layouts for all groups have been incorporated into the scene.

The initial state of each object includes \textbf{object ID, parent object, XY-plane coordinates, and rotation around the Z-axis}, while spatial relationships capture \textbf{object name, facing object, aligned object, related wall, neighboring objects, and distance} (which can be \texttt{none}). These spatial relationships are subsequently used in scene layout optimization. The \textbf{parent object} denotes the supporting plane of the object, which can be the floor, wall, ceiling, or any object ID within the scene. 
Each support plane is treated as a level, objects on the same support plane are regarded as objects at the same level, and the object to which the support plane belongs (the object that owns the support plane, or floor, wall, ceiling) is regarded as the parent of the objects. The hierarchical relationships between objects will later be used for scene layout optimization.
Given the parent, the VLM predicts only the XY-plane coordinates; the Z-axis coordinate is automatically determined based on the parent and the object’s dimensions. If the parent is the floor or another object, the object is physically supported, and its bottom surface aligns with the top surface of the parent. If the parent is a ceiling or wall, the object is suspended, and the Z-axis coordinate corresponds to the hanging position.

After generating the initial layout, the module performs \textbf{object retrieval}. We first encode all objects in the database using an object encoding model (\textbf{SBERT}\cite{reimers2019sentence} is used in this work) to construct an asset vector library. For each object, a feature vector is extracted from its textual description using the object encoding model, and the top 60 most similar object candidates are retrieved from the asset vector library. Next, \textbf{OpenCLIP}~\cite{ilharco2021openclip} is used to compute text-image similarity for each candidate. Candidate similarity is further refined based on the object’s size relative to the target. A final score is computed according to the following formula:
\begin{equation}
\begin{split}
Score_{Final}(i) & = w_{1} \cdot S_{sbert}(i) + w_{2} \cdot S_{clip}(i) \\
                 & \quad + w_{3} \cdot S_{size}(i)
\end{split}
\end{equation}
The highest-scoring object is selected as the retrieved object. Finally, the predicted object dimensions in the scene layout are replaced with the retrieved object’s actual dimensions, producing the final initial scene layout. \textcolor{r4color}{Our framework employs a modular design, allowing for the replacement of retrieval-based asset acquisition methods with generative approaches. More details and results could be found in the supplementary materials.}

\subsection{Hierarchical Optimization}
After each group of objects is placed, an initial scene layout is obtained. Based on this initial layout, the scene optimization module will optimize and adjust the existing layout using the object spatial constraint information predicted by the VLM. For the optimization algorithm, we propose a hierarchical force-directed layout optimizer based on constraints. The core idea of the algorithm is to uniformly abstract and transform all complex scene layout rules—including physical feasibility (e.g., collisions, boundaries) and semantic logic (e.g., proximity, against walls)—into simulated continuous physical forces. These forces, serving as penalty terms, drive the scene layout in an iterative loop from a potentially conflict-ridden initial state to a stable, force-balanced state in which all constraints are satisfied. \textcolor{r4color}{Some examples of the initial and the optimized scene layout could be found in the supplementary materials.}

\subsubsection{System State Definition}
At any discrete time step $t$, the system state of the entire scene layout $\mathcal{S}(t)$ is defined as the set of all $N$ object states:
\begin{equation}
\label{eq::scene}
\mathcal{S}(t) = \{ O_1(t), O_2(t), \dots, O_N(t) \},
\end{equation}
\begin{equation}
\label{eq::state}
O_i(t) = \{ p_i(t), \theta_i(t), s_i(t)\},
\end{equation}
We decompose the position $p_i \in \mathbb{R}^3$ into a planar coordinate $p_{i, \text{plane}} \in \mathbb{R}^2$ ($(X, Y)$) and a vertical coordinate $p_{i, \text{vert}} \in \mathbb{R}$ ($(Z)$). The rotation $\theta_i$ primarily refers to its Yaw, and the scale $s_i\in \mathbb{R}^3$.
\subsubsection{Iterative Optimization Process}
The optimization is a discrete iterative update process $S(t) \to S(t+1)$. At time $t$, the system maintains two independent force accumulators for $O_i(t)$: (1) \textbf{Planar Force Accumulator $F_{i, \text{plane}} \in \mathbb{R}^2$}, which is used to update the object's $(X, Y)$ coordinate; (2) \textbf{Vertical Force Accumulator $F_{i, \text{vert}} \in \mathbb{R}$}, which is used to update the object's $Z$ coordinate. (3) \textbf{Rotational torque accumulator} $\tau_i \in \mathbb{R}$, which is used to update the object's yaw angle.

Optimization requires consideration of various constraints, which are divided into \textbf{physical constraints} and \textbf{semantic constraints}. \textbf{Physical constraints} include collision constraints, boundary constraints, and support constraints, while \textbf{semantic constraints} include proximity constraints, wall-adjacent constraints, orientation constraints, and alignment constraints. For each constraint, we calculate all $k$ force vectors $v_{1...k}$ separately for each object $O_i$. 

\textbf{Horizontal direction} includes collision constraints between objects at the same level, boundary constraints of the room in the XY plane, proximity constraints, support constraints, and wall-adjacent constraints. For collision constraints between objects at the same level, $v_k$ is calculated using the overlapping area of their bounding boxes projected onto the XY plane. For proximity constraints, $v_k$ is calculated using the difference between the distance between the two objects and the target distance. For support constraints, $v_k$ is calculated using the overlapping area of the object's bounding box projection onto the XY plane on the parent-level plane. For wall-adjacent constraints, $v_k$ is calculated using the nearest distance from the object's 2D bounding box to the target wall. For boundary constraints of the room in the XY plane, $v_k$ is calculated using the area of the object's 2D bounding box extending beyond the room's area. All horizontal force vectors $v_k$ are 2D vectors $v_k=(v_x,v_y)$. Finally, the force of a constraint is $\Delta F_{i, \text{plane}}(k)= w_k*v_k$, $w_k$ is weight.

\textbf{Vertical direction} includes collision constraints between objects at different levels, and upper and lower boundary constraints within the room's boundary constraints. For collision constraints between objects at different levels, we calculate $v_k$ by the overlap distance of the projections of two objects at different levels with overlapping XZ and YZ plane bounding boxes on the Z-axis. For the upper and lower boundary constraints of the room, we calculate $v_k$ by the distance by which the projection of the object's bounding box on the Z-axis extends beyond the upper and lower boundaries of the room. All kinds of vertical force vector $v_k$ is a 1D vector $v_k=v_z$. Finally, the force of a constraint is $\Delta F_{i, \text{vert}}(k)= w_k*v_k$, $w_k$ is weight.

\textbf{Rotation direction} includes orientation constraints and alignment constraints. For these two types of rotation constraints, we calculate $\tau_k$ using the difference between the object's current angle and the target angle. This value is applied to make the object rotate toward the target.

After all constraints are evaluated, the total forces for object $O_i$ are given by:
\vspace{-0.3cm}
\begin{equation}
\label{eq::f_plane}
F_{i, \text{plane}}(t) = \sum_{k=1}^{K} \Delta F_{i, \text{plane}}(k), 
\end{equation}
\begin{equation}
\label{eq::f_vertical}
F_{i, \text{vert}}(t) = \sum_{k=1}^{K} \Delta F_{i, \text{vert}}(k),
\end{equation}
\begin{equation}
\label{eq::f_rotation}
\tau_i(t) = \sum_{k=1}^{K} \tau_k,
\end{equation}
\subsubsection{Deadlock Detection and Evasion}

``Deadlock" refers to a situation where an object in a system is  ``stuck" due to equal but opposite forces, preventing it from moving to a better position (i.e., trapped in a local minimum). A deadlock is determined when the cumulative sum of the object's movement distance within a time window in the horizontal or vertical direction exceeds a threshold $\mathcal{D}_1$, and the absolute distance the object moves within the start and end times of the time window is less than a threshold $\mathcal{D}_2$. Deadlocks are categorized as horizontal or vertical deadlock. If a horizontal deadlock is detected, a force perpendicular to the deadlock direction will be applied to ``push" the object out of its stuck state. If a vertical deadlock (e.g., the object is being squeezed vertically), the object's scaling $s'_i(t)$ is directly modified, shrinking it along the Z-axis, thus breaking the deadlock.

\subsubsection{State Update}

The system state is updated based on the modified forces using an Explicit Euler integration step, where the $\alpha$ coefficients are step sizes. 
The planar and vertical positions are updated independently according to:
\begin{equation}
\label{eq::p_plane}
p_{i, \text{plane}}(t+1) = p_{i, \text{plane}}(t) + \alpha_{\text{trans}} \cdot F_{i, \text{plane}}(t),
\end{equation}
\begin{equation}
\label{eq::p_vertical}
p_{i, \text{vert}}(t+1) = p_{i, \text{vert}}(t) + \alpha_{\text{vert}} \cdot F_{i, \text{vert}}(t),
\end{equation}
The rotation is updated by:
\vspace{-0.2cm}
\begin{equation}
\label{eq::rotation_itaration}
\theta_i(t+1) = \theta_i(t) + \alpha_{\text{rot}} \cdot \tau_i(t),
\end{equation}
Finally, the scale is updated, $s_i(t+1) = s'_i(t)$, and this update is triggered only during a vertical deadlock event.

\subsubsection{Convergence Condition}

The iterative process continues until the maximum iteration count $I_{\text{max}}$ is reached, or the system achieves a ``converged" state. Convergence is defined as the total sum of all active constraint forces in the system being less than a small threshold $\epsilon_{\text{conv}}$:
\vspace{-0.3cm}
\begin{equation}
\label{convergence}
\begin{split}
    F_{residual}(t) = & \sum_{i=1}^{N} \left( \|F'_{i, \text{plane}}(t)\| \right. \\ 
                    & \left. + |F'_{i, \text{vert}}(t)| + |\tau_i(t)| \right) 
\end{split}
\end{equation}
\begin{equation}
F_{residual}(t) < \epsilon_{\text{conv}}
\end{equation}
When this condition is met, the system is considered to have reached a stable state.
Then we get the current final scene.
The optimization module will then perform the next round of scene updates, including the next group, until all objects are added to the scene, resulting in the final scene.

\begin{table*}[t]
\renewcommand{\arraystretch}{0.5}
  \caption{Evaluation results of different methods.}
  \vspace{-0.2cm}
\centering
   \resizebox{\linewidth}{!}
   {
    \begin{tabular}{@{} l rrrrrrrrrrrrr @{}} 
    \toprule
    \toprule
        & \multicolumn{4}{c}{Fidelity} 
        & \multicolumn{6}{c}{Plausibility} 
        & 
        & 
        & 
        \\ 
    \cmidrule(l){2-5} \cmidrule(l){6-11} 
   
        & $\uparrow~$CNT$_\%$ & $\uparrow~$ATR$_\%$ & $\uparrow~$OOR$_\%$ & $\uparrow~$OAR$_\%$ 
        & $\downarrow~$COL$_{ob}\%$ & $\downarrow~$COL$_{sc}\%$ & $\uparrow~$SUP$_\%$ & $\uparrow~$NAV$_\%$ & $\uparrow~$ACC$_\%$ & $\downarrow~$OOB$_\%$ 
        & $\downarrow~$Time$(s)$      
        & $\uparrow~$SP       
        & $\uparrow~$CLIPsim     \\
    \midrule 
   
    LayoutGPT & 29.02        & 22.48        & 6.11        & 11.48        & 35.67         & 49.00         & 34.39         & \textbf{99.52}   & 73.77        & 35.81        & \textbf{37.41}    & 35.14     & 16.22      \\
    Holodeck  & 39.98        & 45.62        & 18.99        & 38.27        & 12.24         & 63.00         & 34.72        & 99.29        & 91.21        & \textbf{0.34}    & 271.97        & 55.45     & 18.25      \\
    LayoutVLM & 76.72        & \textbf{61.79}   & 38.09        & 61.99        & 29.44         & 55.00         & 77.54        & 96.96        & 87.67        & 8.42          & 321.47        & 65.54      & 18.40      \\ 
    \midrule 
    HOG-Layout & \textbf{77.84}   & 60.01        & \textbf{43.09}   & \textbf{75.74}   & \textbf{5.28}     & \textbf{16.00}     & \textbf{81.17}   & 97.15        & \textbf{94.74}   & 2.45          & 70.44         & \textbf{69.69} & \textbf{18.61} \\
    \bottomrule
    \bottomrule
    \end{tabular}
   }
\vspace{-0.4cm}
   \label{tab::results}
  \end{table*}

\begin{figure}[h]
  \centering
   \includegraphics[width=\linewidth]{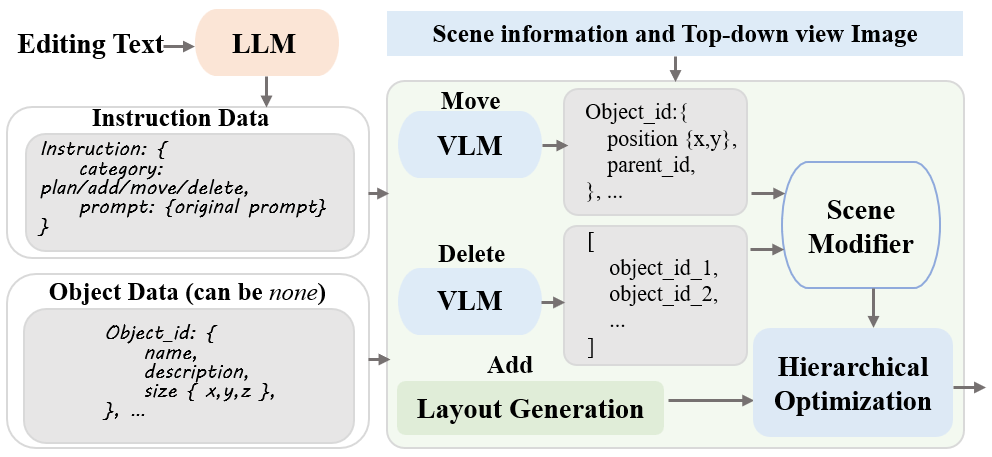}
   \caption{The editing pipeline of HOG-Layout.}
   \label{fig:editing}
\end{figure}

\subsection{Scene Editing}
Beyond generating complete scene layouts, HOG-Layout also supports editing scenes. As in \Cref{fig:editing} when input a text, the text instruction is routed by the LLM; any user text instruction will be mapped to one of four basic commands: plan, add, delete, or move. Among these, ``plan'' represents scene generation, where object data is \textit{none}. When the text is determined to be one of the other three editing commands, a new prompt is constructed from the text, object data, and scene information. If ``add", the prompt and top-down view image are input to the \textbf{Layout Generation} module, and the subsequent process is the same as in scene generation process. If ``move", the prompt and top-down view image are input to the VLM, which outputs the IDs of all objects to be moved and their new positions, and then this text information is input to the \textbf{scene modifier}. If ``delete", the prompt and top-down view image are input to the VLM, which outputs the IDs of all deleted objects, and then this text information is input to the \textbf{scene modifier}. The \textbf{scene modifier} retrieves the corresponding object information from the original scene layout, performs the modifications, generates a new layout, and then inputs it to  \textbf{Hierarchical Optimization} module to modify the scene.

\section{Experiments}
\label{sec::exp}

We conduct comparative and ablation experiments on the SceneEval benchmark\cite{tam2026sceneeval}. For different methods, we uniformly use the GPT-4o\cite{achiam2023gpt} model for scene generation with 3D assets sourced from 3D-FUTURE\cite{fu20213d} and Objaverse\cite{deitke2023objaverse}. Additionally, we design 30 scene editing prompts to evaluate scene editing ability, based on 10 generated scenes. For each scene, we design three types of instructions: adding, deleting, and moving objects. There are two types of scenes: 5 are \textbf{simple scenes} where the total number of objects is fewer than 5, and 5 are \textbf{complex scenes} where the total number of objects is greater than 7. 
\subsection{Dataset}
\textbf{SceneEval}~\cite{tam2026sceneeval} is a benchmark designed for evaluating text-conditioned 3D indoor scene generation. We adopt SceneEval-100 dataset, which contains text descriptions of 100 indoor scenes, covering 10 different room types. These descriptions are categorized into three difficulty levels—easy, medium, and hard—based on their complexity.

\subsection{Metrics}

We follow the SceneEval to evaluate scene quality. Unlike traditional metrics that focus only on distributional similarity, SceneEval divides metrics into two major categories: \textbf{Fidelity} and \textbf{Plausibility}. \textbf{Fidelity} metrics measure the consistency between the generated scene and the input text description, including: \textbf{CNT} (Object Count), \textbf{ATR} (Object Attribute), \textbf{OOR} (Object-Object Relationship), and \textbf{OAR} (Object-Architecture Relationship). \textbf{Plausibility} metrics evaluate whether the scene conforms to physical common sense, including: \textbf{COL} (Object Collision), \textbf{SUP} (Object Support), \textbf{NAV} (Scene Navigability), \textbf{ACC} (Object Accessibility), and \textbf{OOB} (Object Out-of-Bounds). Additionally, we compare the average \textbf{generation time} for 100 scenes produced by different methods, and use the LONG-CLIP-L model\cite{zhang2024long} to test the \textbf{CLIPsim} (CLIP similarity) between the generated scenes and the text prompts. Finally, we use GPT-5\cite{singh2025openai} to score the \textbf{SP} (semantic plausibility, ranging from 0 to 100) of the 100 scenes generated by each method.

\subsection{Compared Methods}

We compare our method with LayoutGPT~\cite{feng2023layoutgpt}, Holodeck~\cite{yang2024holodeck}, and LayoutVLM~\cite{sun2025layoutvlm}. Since LayoutVLM optimizes layout based on text instructions and a resource list, we use the text instructions and the object list obtained by our method as the input to LayoutVLM, and generate the final scene layout. To evaluate \textbf{CLIPsim}, even though our method supports the generation of non-specific objects, we still used text prompts containing detailed object information for generation.

\subsection{Results}

\begin{figure}[htbp]
\centering
\includegraphics[width=0.46\textwidth]{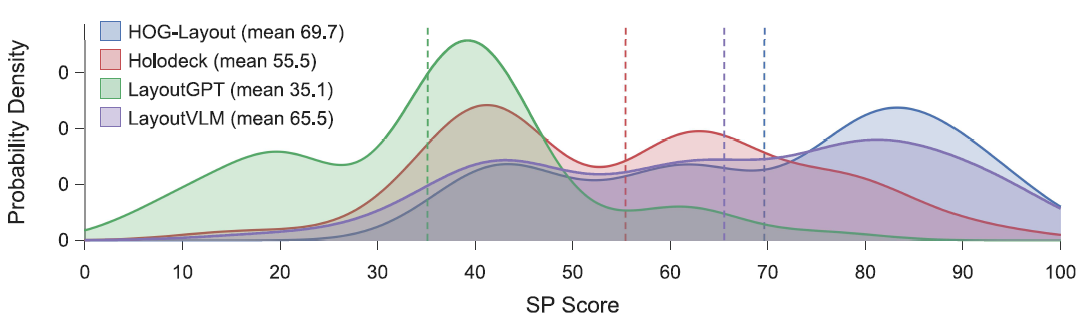} 
\caption{Results of SP Score Distribution (Kernel Density Estimation, KDE) of  different methods.}
\label{fig::sp}
\end{figure}

\begin{figure*}
    \centering
    \includegraphics[width=0.85\textwidth]{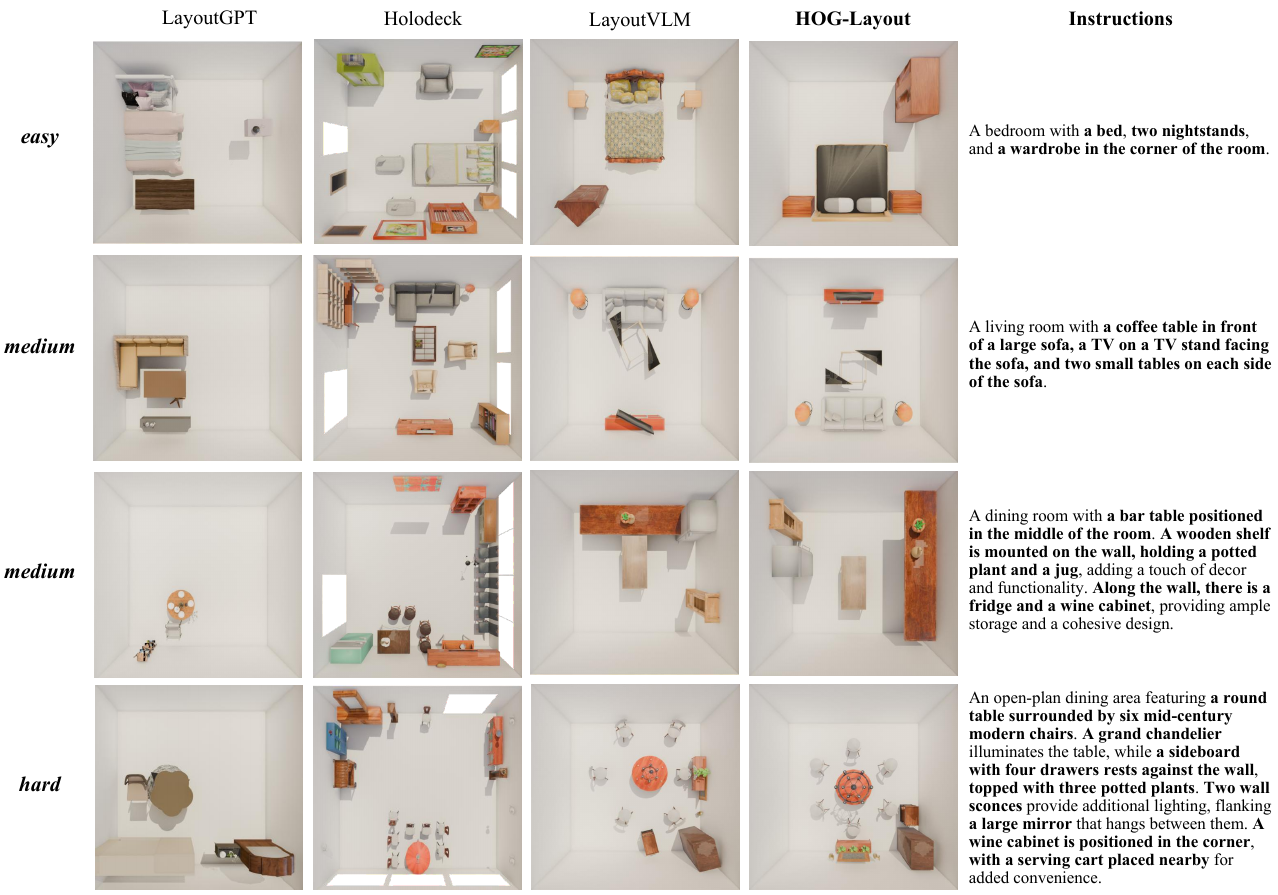}
    \caption{Generation Examples of different methods on the benchmark. 
    }
    \label{fig:qualitative}
\end{figure*}

\Cref{fig:qualitative} shows some examples of different methods. From the visualization results, our method produces more realistic, accurate scenes that better match the textual instructions. In the Medium-level Living Room scene of \Cref{fig:qualitative}, only our method fully satisfies the instructions and places the coffee table in the correct position and orientation. Observing these results, we find that LayoutGPT often generates physically implausible scenes. While Holodeck can produce reasonable scenes, it frequently violates the object categories and spatial relationships specified in the instructions. LayoutVLM can follow user instructions to generate reasonable and instruction-compliant scenes, yet the positions and rotations of objects are not sufficiently accurate and sometimes contain flaws. Only our method can generate reasonable and accurate scene layouts while fully complying with the instructions.

\Cref{tab::results} shows the evaluation results and \Cref{fig::sp} shows the SP score distribution. Our method outperforms other baselines in fidelity, plausibility, CLIP similarity, and semantic plausibility scores. LayoutGPT performs best in generation time and scene navigability, but lags in fidelity, physical plausibility, and other metrics. This is likely because the layout is generated directly from LLM without any optimization steps. 
Holodeck achieves the best OOB score; its layouts computed with DFS or MILP solvers satisfy hard physical constraints well, yet fail to meet soft semantic constraints. LayoutVLM incorporates our retrieval method, so its CNT and ATR results are comparable to ours, and it surpasses Holodeck and LayoutGPT on most other metrics, but its average scene-generation time is excessively long due to time-consuming gradient-based optimization. Compared with LayoutVLM, our method produces more effective and accurate scenes in significantly less time.  

\subsection{Human evaluation}
\textcolor{r4color}{We conducted a user study following the work~\cite{sun2025layoutvlm}. We recruited 15 participants to rate 10 generated scenes per method on a 7-point scale. Participants were given the same instructions used as prompts for the GPT-5 evaluator. \Cref{tab:human_eval} shows the results, which are consistent with GPT-based evaluation as in the work~\cite{sun2025layoutvlm}.}

\begin{table}[htbp]
\renewcommand{\arraystretch}{0.5}
  \centering
  \caption{Human evaluation results.}
  \vspace{-0.2cm}
  \resizebox{0.8\linewidth}{!}
  {\begin{tabular}{@{}lcc@{}}
    \toprule
    \toprule
    Method & Plausibility $\uparrow$ & Semantic Alignment $\uparrow$ \\
    \midrule
    LayoutGPT & 2.43 ($\pm$ 0.54) & 2.58 ($\pm$ 0.62) \\
    Holodeck & 3.97 ($\pm$ 0.91) & 3.66 ($\pm$ 0.76) \\
    LayoutVLM & 3.69 ($\pm$ 0.68) & 4.61 ($\pm$ 0.92) \\
    \textbf{HOG-Layout} & \textbf{5.33 ($\pm$ 0.88)} & \textbf{5.75 ($\pm$ 0.87)} \\
    \bottomrule
    \bottomrule
  \end{tabular}
  }
  \label{tab:human_eval}
\end{table}

\begin{figure}[t]
\centering
\includegraphics[trim={0cm 0cm 0cm 0.5cm}, clip,width=0.4\textwidth]{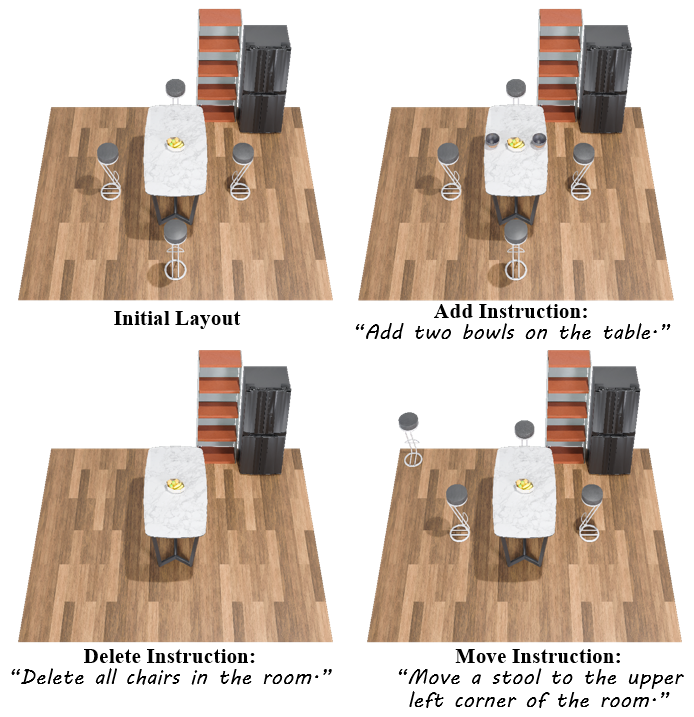} 
\caption{Editing Examples.} 
\label{fig::edit}
\end{figure}

\subsection{Ablation Study}
\begin{table*}[htbp]
\renewcommand{\arraystretch}{0.5}
  \centering
  \caption{Results of ablation study.}
  \vspace{-0.2cm}
  \label{tab:result2}
  \resizebox{\linewidth}{!}
  {
  \begin{tabular}{@{} l rrrrrrrrrrr @{}} 
  \toprule
  \toprule
                & \multicolumn{2}{c}{Fidelity}                          & \multicolumn{6}{c}{Plausibility}                                                                                                                                                         &            &                   &           \\ 
                
  \cmidrule(lr){2-3} \cmidrule(lr){4-9}
  
                & $\uparrow~$OOR$_\%$       & $\uparrow~$OAR$_\%$       & $\downarrow~$COL$_{ob}\%$   & $\downarrow~$COL$_{sc}\%$   & $\uparrow~$SUP$_\%$       & $\uparrow~$NAV$_\%$                & $\uparrow~$ACC$_\%$                & $\downarrow~$OOB$_\%$     &$\downarrow~$Time$(s)$ & $\uparrow~$SP           & $\uparrow~$CLIPsim   \\ 
   \midrule 
  
  w/o~ RAG                  & 37.89                     & 68.98                     & 6.24                      & 20.00                     & 83.34            & 96.56                              & 93.42                              & 4.84                      & 82.26                              & 67.08                     & 18.46     \\
  w/o~ layout description   & 40.42                     & 67.14                     & 7.54                      & 16.00                     & \textbf{83.48}                     & 97.33                              & 94.34                              & 4.55                      & 75.87                              & 67.58                     & 18.95     \\
  w/o~ object data          & 38.23                     & 65.07                     & 7.80                      & 22.00                     & 82.16                     & 97.83                              & 94.46                              & 6.61                      & 76.09                              & 65.61                     & \textbf{18.95}     \\
  only room planning             & 39.06                     & 67.63                     & 6.26                      & 19.00                     & 82.52                     & 97.70                              & 94.91                              & 2.70                      & 67.66                              & 68.02                     & 18.43     \\
  w/o~ grid                 & 41.25                     & 67.60                     & 6.53                      & 19.00                     & 83.07                     & 97.84                              & \textbf{95.61}                     & 5.54                      & \textbf{66.82}                     & 67.51                     & 18.53     \\
  w/o HierOpt               & 35.85                     & 62.55                     & 36.46                     & 65.00                     & 72.65                     & \textbf{98.69}                     & 82.68                              & 19.75                     & 71.97                              & 66.20                     & 17.81     \\
  
  w/ GradOpt                & 42.21                     & 71.37                     & 11.64                     & 28.00                     & 80.65                     & 97.69                              & 92.32                              & 2.94                      & 147.93                             & 67.82                     & 18.53     \\ 
  \midrule 
  HOG-Layout                & \textbf{43.09}            & \textbf{75.74}            & \textbf{5.28}             & \textbf{16.00}            & 81.17                     & 97.15                              & 94.74                              & \textbf{2.45}             & 70.44                              & \textbf{69.69}            & 18.61    \\
  \bottomrule
  \bottomrule
  \end{tabular}
  }
  \vspace{-0.5cm}
  \label{tab::ablation}
  \end{table*}

\begin{table}[htbp]
\renewcommand{\arraystretch}{0.5}
  \caption{Comparison of retrieval methods.}
  \vspace{-0.2cm}
  \centering
  \resizebox{0.65\linewidth}{!}
  {\begin{tabular}{lll} 
  \toprule
  \toprule
  Retrieval Methods & $\uparrow~$CNT$_\%$            & $\uparrow~$ATR$_\%$             \\ 
  \midrule 
  CLIP                                    & 75.34          & 63.05           \\
  CLIP+SBERT                              & 76.30          & 61.13           \\
  CLIP+SBERT+SIZE                       & \textbf{77.86} & \textbf{64.21}  \\
  \bottomrule
  \bottomrule
  \end{tabular}}
  \label{tab::retrieval}
  \end{table}

\textcolor{r4color}{In w/o layout description, we remove the layout description from the planning module, while the rest of the planning module is kept.
In w/o object data, we remove the information from the object data within the planning module.
As room planning contains basic room information and is essential for subsequent rendering and prediction processes, in only room planning, we retain only the room planning part of the planning module, while the rest of the planning module is removed.}
In w/o grid, we remove the grid lines from the image fed to the VLM. 
In w/o HierOpt, we remove the scene optimization module. 
In w/ GradOpt, we use the gradient-descent optimization algorithm from LayoutVLM in our method for comparison.

As shown in \Cref{tab::ablation}, the results verify the effectiveness of our design. Although the scene planning module and its internal RAG module do not directly participate in the VLM’s scene prediction process, the scene plan and object placement guidance they produce, when embedded in the prompt, effectively improve the VLM’s prediction quality. 
\textcolor{r4color}{The results of only room planning, w/o layout description, and w/o object data indicate that every part of the planning module is effective, with the object data part having a more significant impact on the quality of scene generation.}
The results of w/o HierOpt and w/ GradOpt demonstrate that the hierarchical iterative optimizer greatly boosts scene generation quality, and compared with the gradient-based optimizer, our method requires significantly less time for scene generation while achieving better optimization. 

In addition, we evaluate the effectiveness of each component in the retrieval method since \textbf{CNT} and \textbf{ATR} are related to retrieval methods. We compare with methods only using CLIP similarity, the sum of CLIP and SBERT similarities, and the method that includes size. For the same object list, we use different methods to retrieve objects and generate scenes, then compare the CNT and ATR metrics. As shown in \Cref{tab::retrieval}, the results show that combining CLIP, SBERT, and size improves the accuracy of retrieval. This step can directly use text descriptions to generate virtual objects in the future, replacing the current retrieval process.

\subsection{Evaluation on Scene Editing}
\begin{table}[htbp]
\renewcommand{\arraystretch}{0.8}
\centering
\caption{Editing time results.}
\vspace{-0.2cm}
\resizebox{0.8\linewidth}{!}
{
\begin{tabular}{lll} 
\toprule
\toprule
\multicolumn{1}{l}{Instruction type} & Time$_{simple}$  & TIME$_{complex}$   \\ 
\hline
Move                               & $15.64 \pm 1.55$ & $16.35 \pm 3.58$  \\
Delete                             & $15.55 \pm 1.14$  & $18.99 \pm 11.56$ \\
Add                                & $18.39 \pm 1.38$ & $23.93 \pm 6.15$ \\
\bottomrule
\bottomrule
\end{tabular}
}
\label{tab::edit}
\end{table}
As shown in \Cref{tab::edit}, we measure the completion time for each editing operation. 
Except for the add-object in complex scenes, the average time for all other editing doesn't exceed 20 seconds. This demonstrates that our method supports real-time interaction.
The time required in complex scenes is greater than that in simple scenes, which means that the number of objects in the scene affects the speed in editing. Among the three different types of instructions, the add-object instruction takes the longest time; when adding objects, it is necessary to build an object list by the large language model and retrieve assets from the asset library, while also considering the reasonableness of item placement, thus consuming more time than other instructions.

\section{Conclusion}

In this paper, we presented \textbf{HOG-Layout}, a hierarchical scene generation and optimization framework that integrates an LLM and a VLM for text-driven 3D scene synthesis and editing. By introducing a hierarchical representation of spatial support relationships, our method effectively decomposes the reasoning space, enabling VLMs and optimization module to focus on inter-object dependencies within the same or parent-child layers. This design significantly improves both the physical plausibility and computational efficiency of layout reasoning. Moreover, by leveraging a template-based rule library and retrieval-augmented generation, our approach ensures that the generated scenes are semantically consistent and spatially reasonable. Experimental results demonstrate that HOG-Layout achieves fast and coherent scene generation while supporting intuitive and low-latency text-based editing.
We hope our work will inspire future research in 3D scene generation and editing, and further empower embodied intelligence and immersive virtual environments.

\section*{Acknowledgement}
This work was supported by the New Generation Artificial Intelligence-National Science and Technology Major Project (2025ZD0124000), the National Natural Science Foundation of China under Grants (No.62502415), and the Hong Kong Polytechnic University under Projects No.P0046486, No.P0046226, and No.P0053828.
{
    \small
    \bibliographystyle{ieeenat_fullname}
    \bibliography{main}
}
\clearpage
\setcounter{page}{1}
\maketitlesupplementary
\appendix
\definecolor{r4color}{RGB}{0, 0, 0}

\section{Details of the Experiment Setup}
In the comparative experiment and ablation study, for all VLM and LLM usage, we use GPT-4o-2024-08-06 with the default parameters. The experiment is conducted on a single computer with the following specifications:
\begin{itemize}
    \item \textbf{OS}: Windows 11
    \item \textbf{CPU}: 13th Gen Intel Core i5-13500H
    \item \textbf{GPU}: NVIDIA GeForce RTX 4060 (8GB)
    \item \textbf{Memory}: 64 GB
\end{itemize}

\section{Details of \textbf{HOG-Layout}}

\subsection{Optimization Algorithm Details}
We model layout optimization as a physics simulation (\cref{alg:main_loop}). To maintain notational clarity, we denote the force calculation functions in the pseudocode using capitalized abbreviations (e.g., $COLH$ for Horizontal Collision), which directly correspond to the constraint details described below. In each iteration $t$, we calculate specific force components corresponding to different constraints:

\begin{algorithm}[t]
    \caption{Hierarchical Force-Directed Optimization}
    \label{alg:main_loop}
    \begin{algorithmic}[1]
    \Require Objects $\mathcal{S} = \{O_1, \dots, O_N\}$, Room $\mathcal{R}$, Params $\Theta$
    \Ensure Optimized $\mathcal{S}_{\text{opt}}$
    
    \State $t \leftarrow 0, \quad converged \leftarrow \text{False}$
    
    \While{$t < T_{\max} \land \neg converged$}
        \State \textbf{Init:} $F_{i, \text{plane}} \leftarrow \mathbf{0}, F_{i, \text{vert}} \leftarrow 0, \tau_i \leftarrow 0, \forall O_i \in \mathcal{S}$
        \State $active \leftarrow \text{False}$
    
        \ForAll{$O_i \in \mathcal{S}$}
            \State \{1. Calculate Different Constraint Forces\}
            \State $F_{\text{bnd}}^{\text{h}} \leftarrow$ \textsc{BndH}($O_i, \mathcal{R}$)
            \State $F_{\text{bnd}}^{\text{v}} \leftarrow$ \textsc{BndV}($O_i, \mathcal{R}$)
            \State $F_{\text{sup}} \leftarrow$ \textsc{Sup}($O_i, O_{\text{parent}}$)
            \State $F_{\text{col}}^{\text{h}} \leftarrow$ \textsc{ColH}($O_i, \mathcal{S}$)
            \State $F_{\text{col}}^{\text{v}} \leftarrow$ \textsc{ColV}($O_i, \mathcal{S}$)
            
            \State $F_{\text{adj}} \leftarrow$ \textsc{Adj}($O_i, O_{\text{target}}$)
            \State $F_{\text{wall}} \leftarrow$ \textsc{Wall}($O_i, \mathcal{R}$)
            \State $\tau_{\text{align}} \leftarrow$ \textsc{Align}($O_i, O_{\text{target}}$)
            \State $\tau_{\text{pnt}} \leftarrow$ \textsc{Point}($O_i, O_{\text{target}}$)
            
            \State \{2. Accumulate All the Forces\}
            \State $F_{i, \text{plane}} \leftarrow F_{i, \text{plane}} + w_{\text{bnd}} F_{\text{bnd}}^{\text{h}} + w_{\text{col}} F_{\text{col}}^{\text{h}} + w_{\text{sup}} F_{\text{sup}} + w_{\text{adj}} F_{\text{adj}} + w_{\text{wall}} F_{\text{wall}}$
            \State $F_{i, \text{vert}} \leftarrow F_{i, \text{vert}} + w_{\text{bnd}} F_{\text{bnd}}^{\text{v}} + w_{\text{vcol}} F_{\text{col}}^{\text{v}}$
            \State $\tau_{i} \leftarrow \tau_{i} + w_{\text{align}}\tau_{\text{align}} + w_{\text{pnt}}\tau_{\text{pnt}}$
            
            \State \{3. Check Convergence\}
            \State $F_{\text{total}} \leftarrow \|F_{i, \text{plane}}\| + |F_{i, \text{vert}}| + |\tau_i|$
            \If{$F_{\text{total}} > \epsilon$} \State $active \leftarrow \text{True}$ \EndIf
        \EndFor
    
        \State \{4. Deadlock \& Update\}
        \State $Deadlock \leftarrow$ \textsc{HandleDeadlocks}($\mathcal{S}, \Theta$)
        
        \ForAll{$O_i \in \mathcal{S}$}
            \State $p_{i, \text{plane}}^{(t+1)} \leftarrow p_{i, \text{plane}}^{(t)} + \eta_{\text{trans}} \cdot F_{i, \text{plane}}$
            \State $p_{i, \text{vert}}^{(t+1)} \leftarrow p_{i, \text{vert}}^{(t)} + \eta_{\text{vert}} \cdot F_{i, \text{vert}}$
            \State $\theta_{i}^{(t+1)} \leftarrow \theta_{i}^{(t)} + \eta_{\text{rot}} \cdot \tau_{i}$
        \EndFor
        
        \If{$\neg active \land \neg Deadlock$} \State $converged \leftarrow \text{True}$ \EndIf
        \State $t \leftarrow t + 1$
    \EndWhile

    \State \textbf{return} $\mathcal{S}$
    \end{algorithmic}
    \end{algorithm}

\begin{algorithm}[t]
\caption{Deadlock Detection and Evasion}
\label{alg:deadlock}
\begin{algorithmic}[1]
\Require Object $O_i$, History $\mathcal{H}_i$, Contributions $\Phi_i$, Params $\Theta$
\Ensure Boolean $is\_deadlocked$
\State $is\_deadlocked \leftarrow \text{False}$

\State \{1. Check Active Evasion Timer\}
\If{$O_i.timer > 0$}
    \State $F_{i, \text{plane}} \leftarrow F_{i, \text{plane}} + F_{i, \text{evade}}$
    \State $O_i.timer \leftarrow O_i.timer - 1$
    \State \textbf{return} $\text{True}$
\EndIf

\State \{2. Horizontal Deadlock\}
\If{$\sum \mathcal{H}_i[t-W:t] < \epsilon_{\text{disp}} \land |\Phi_i| \geq 2$}
    \State Select pair $(\mathbf{v}_a, \mathbf{v}_b) \in \Phi_i$ with \textbf{Max Mag Sum} s.t. $\angle(\mathbf{v}_a, \mathbf{v}_b) \approx 180^\circ$
    
    \If{such pair exists}
        \State $\mathbf{v}_{\text{axis}} \leftarrow \frac{\mathbf{v}_a}{\|\mathbf{v}_a\|}$ \Comment{Normalized Axis}
        \State $\mathbf{v}_{\perp} \leftarrow (-\mathbf{v}_{\text{axis}, y}, \mathbf{v}_{\text{axis}, x})$ \Comment{Orthogonal}
        \State $F_{i, \text{evade}} \leftarrow \lambda_{\text{evade}} \cdot \mathbf{v}_{\perp}$
        \State $O_i.timer \leftarrow T_{\text{deadlock}}$
        \State $F_{i, \text{plane}} \leftarrow F_{i, \text{plane}} + F_{i, \text{evade}}$
        \State $is\_deadlocked \leftarrow \text{True}$
    \EndIf
\EndIf

\State \{3. Vertical Deadlock\}
\If{Vertical displacement small $\land F_{\text{vert}}^{\text{up}} > 0, F_{\text{vert}}^{\text{down}} < 0$}
    \State Calc available gap $h_{\text{gap}}$
    \If{$h_{\text{gap}} < h_i$}
        \State $s_i \leftarrow s_i \cdot (h_{\text{gap}} / h_i)$ \Comment{Adjust Scale}
        \State $is\_deadlocked \leftarrow \text{True}$
    \EndIf
\EndIf

\State \textbf{return} $is\_deadlocked$
\end{algorithmic}
\end{algorithm}

\begin{itemize}
    \item \textbf{Physical Constraints}:
    \begin{itemize}
        \item \textbf{Boundary ($F_{\text{bnd}}^{\text{h}}, F_{\text{bnd}}^{\text{v}}$)}: Computed by projecting the object onto the room's axes (X, Y, Z). If the projection interval lies outside the room boundaries defined by $\mathcal{R}$, a restoring force proportional to the penetration depth is applied to push the object back.
        \item \textbf{Support ($F_{\text{sup}}$)}: Ensures stability by checking the contact surface area between the object and its parent. We calculate the intersection polygon of their footprints; if the ratio of the intersection area to the object's area is below a threshold, a centripetal force is applied to pull the object towards the parent's geometric center.
        \item \textbf{Horizontal Collision ($F_{\text{col}}^{\text{h}}$)}: $F_{\text{col}}^{\text{h}}$ resolves collisions between same-level objects. The HOG-Layout framework supports two distinct approaches for resolving overlaps, both of which are viable depending on the precision requirements:
        \begin{enumerate}
            \item \textit{Area-based Method}: A simpler approach where the force magnitude is proportional to the overlapping area of the bounding boxes, and the direction is determined by the vector connecting the centroids.
            \item \textit{SAT-based Method}: A more precise approach (employed in our final implementation) using the \textit{Separating Axis Theorem (SAT)}. It calculates the \textit{Minimum Translation Vector (MTV)} required to separate polygons, providing exact direction and magnitude for the collision force.
        \end{enumerate}
        \item \textbf{Vertical Collision ($F_{\text{col}}^{\text{v}}$)}: It is used for objects at different hierarchy levels (e.g., wall-mounted pictures vs. objects on floor). It is heuristic: for objects between different levels, we first check if the 2D bounding boxes overlap. If they do, we calculate the overlap on the Z-axis. Repulsive forces are then applied strictly along the Z-axis—pushing the higher object upwards and the lower object downwards—to separate them.
    \end{itemize}
    
    \item \textbf{Semantic Constraints}:
    \begin{itemize}
        \item \textbf{Adjacent ($F_{\text{adj}}$)}:This is modeled as a damped spring system. We compute the Euclidean distance between the nearest points of two objects. The force magnitude is proportional to the difference $(d_{current} - d_{target})$, acting along the vector connecting these points to either attract or repel the object.
        \item \textbf{Against Wall ($F_{\text{wall}}$)}: It identifies the nearest wall specified by the instruction (e.g., "left", "back"). A linear force penalizes the perpendicular distance from the object's bounding box edge to the wall plane if it exceeds a tolerance threshold.
        \item \textbf{Alignment \& Pointing ($\tau_{\text{align}}, \tau_{\text{pnt}}$)}: Unlike linear forces, these generate rotational torque. We calculate the shortest angular difference $\Delta \theta$ between the object's current yaw and the target vector (derived from a fixed angle or the relative position of another object). The torque is proportional to $\Delta \theta$ to iteratively correct orientation.
    \end{itemize}
\end{itemize}

\begin{table}[t]
\centering
\caption{Optimized Parameters for the Hierarchical Force-Directed Optimizer.}
\label{tab:hyperparams}
\begin{tabular}{l l c}
\toprule
\textbf{Symbol} & \textbf{Description} & \textbf{Value} \\
\midrule
$w_{\text{col}}$ & Collision weight (Horizontal) & 1.134 \\
$w_{\text{vcol}}$ &  Collision weight (Vertical) & 2.850 \\
$w_{\text{bnd}}$ & Boundary weight & 2.857 \\
$w_{\text{sup}}$ & Support weight & 0.525 \\
$w_{\text{adj}}$ & Adjacent weight & 0.833 \\
$w_{\text{wall}}$ & Against-wall weight & 2.977 \\
$w_{\text{pnt}}$ & Point-towards weight & 4.688 \\
$w_{\text{align}}$ & Alignment weight & 5.037 \\
\midrule
$\lambda_{\text{evade}}$ & Deadlock evasion strength & 0.161 \\
$T_{\text{deadlock}}$ & Deadlock evasion duration (steps) & 17 \\
\midrule
$\eta_{\text{trans}}$ & Translation step size & 0.208 \\
$\eta_{\text{rot}}$ & Rotation step size (degrees/step) & 8.569 \\
$T_{\max}$ & Maximum iterations & 300 \\
\bottomrule
\end{tabular}
\end{table}

These forces are aggregated into planar ($F_{i, \text{plane}}$) and vertical ($F_{i, \text{vert}}$) totals, and the system state is updated via Explicit Euler integration.

\begin{figure}[thbp]
    \centering
    \includegraphics[width=0.96\linewidth]{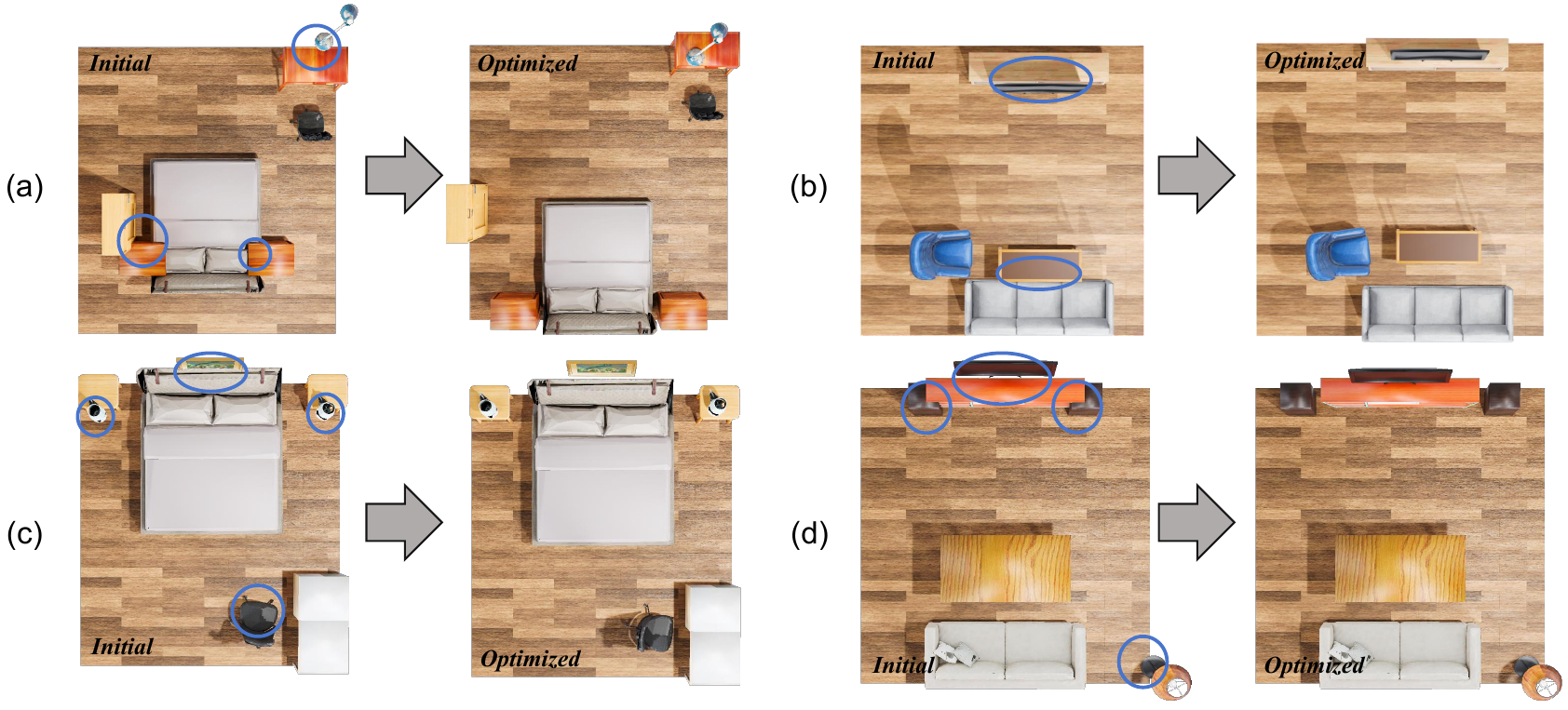}
    \caption{Optimization examples.
    }
    \label{fig:optimize}
\end{figure}

Gradient-free force-directed methods are prone to local minima. To robustly detect deadlocks, we maintain two state variables for each object: a displacement history window $\mathcal{H}_i$ (tracking recent movement magnitudes) and a set of active force contribution vectors $\Phi_i$. \cref{alg:deadlock} addresses this via two mechanisms:
\begin{itemize}
    \item \textbf{Horizontal Deadlock}: A horizontal deadlock is flagged when the cumulative displacement over window $W$ (iteration 20 times) is negligible ($\sum \mathcal{H}_i < \epsilon_{\text{disp}}$), yet the object is subject to multiple competing forces ($|\Phi_i| \ge 2$). If these conditions are met, the algorithm searches $\Phi_i$ for a pair of forces $(\mathbf{v}_a, \mathbf{v}_b)$ that are structurally opposing (angle $\approx 180^\circ$) and have the maximum combined magnitude. 
    It then applies a temporary orthogonal evasion force ($\mathbf{v}_{\perp}$) scaled by $\lambda_{\text{evade}}$ for a duration $T_{\text{deadlock}}$ to break the symmetry.
    \item \textbf{Vertical Deadlock}: Addresses squeezing where an object cannot fit vertically. If the available vertical gap $h_{\text{gap}}$ is smaller than the object height $h_i$, the object scale $s_i$ is adjusted.
\end{itemize}

\subsection{Implementation Details \& Parameters}

\begin{figure*}[t] 
    \centering
    \begin{subfigure}[b]{0.75\textwidth} 
        \centering
        \includegraphics[width=\textwidth]{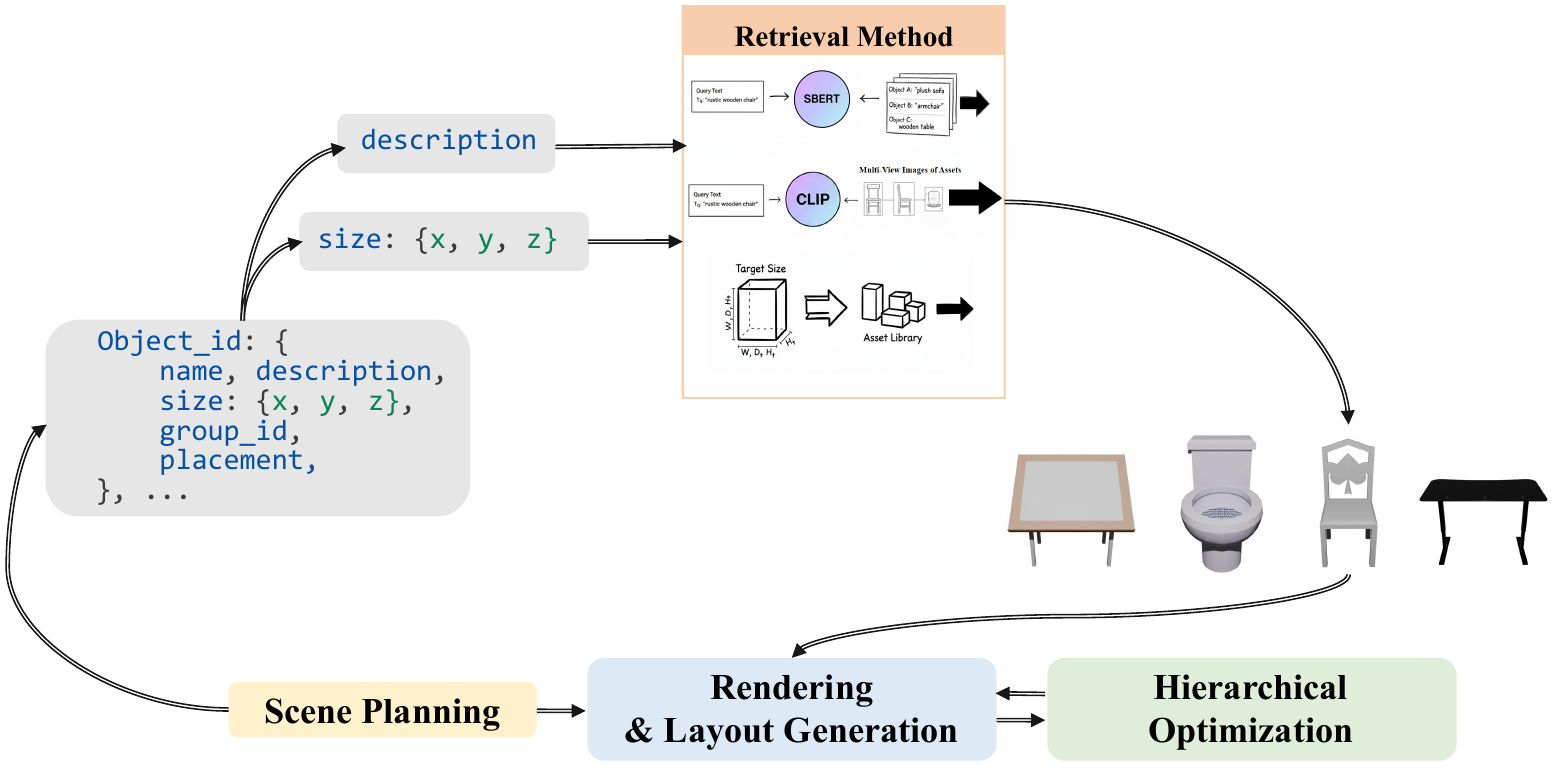} 
        \caption{Pipeline with retrieved objects}
        \label{fig:retrieval_pipeline}
    \end{subfigure}
    \hfill
    \begin{subfigure}[b]{0.75\textwidth} 
        \centering
        \includegraphics[width=\textwidth]{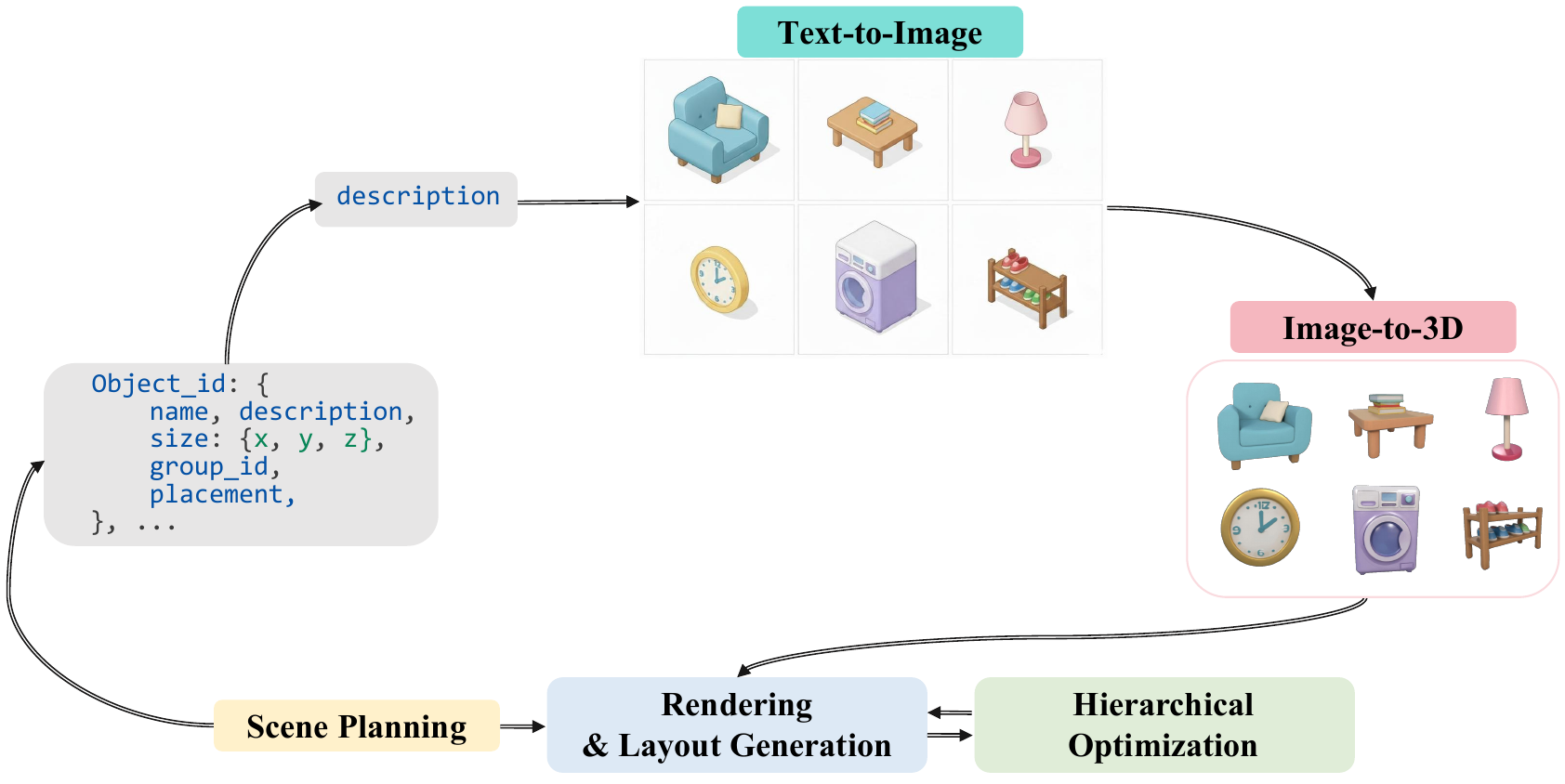}
        \caption{Pipeline with generative 3D objects}
        \label{fig:generative_pipeline}
    \end{subfigure}
    \caption{\textbf{Optional object acquisition methods.} (a) The retrieval-based pipeline matches query text and geometry against a fixed database using semantic and visual similarity. (b) The generative pipeline replaces retrieval with generative objects: generating an image from text (e.g., via DALL-E) and then converting it to a 3D model (e.g., via Hunyuan 3D).}
    \label{fig:pipeline_comparison}
\end{figure*}

\textcolor{r4color}{To ensure stability and convergence of the physics-based simulation, appropriate simulation parameters ($\Theta$) are critical. We \textbf{treat them as hyperparameters} for the purpose of optimization and employ automated tuning techniques to find their optimal values.}

\begin{itemize}
    \item \textcolor{r4color}{\textbf{Necessity of Hyper-parameter Search:} Our iterative optimization process mimics gradient descent, where the various ``forces'' act as gradients, and their corresponding weights function similarly to learning rates. If the weight values are too small, objects move extremely slowly (analogous to vanishing gradients), failing to reach a stable state within the maximum iteration limit. Conversely, if the weights are too large, objects experience severe positional oscillations in each step (analogous to exploding gradients) and can never settle into a balanced position. Therefore, a hyper-parameter search is essential to find a reasonable set of weights that allows the scene layout to converge more quickly and smoothly to a stable state.}
    
    \item \textcolor{r4color}{\textbf{Search Objective and Dataset:} The objective of this search is to minimize a composite penalty. We optimize the parameters on a custom dataset consisting of 50 diverse initial scene layouts generated by our pipeline with GPT-4o (distinct from the SceneEval benchmark). For each scene, we calculate the sum of constraint violations and the residual force magnitudes (which indicate the severity of the violations) at the end of the optimization. The search aims to minimize the average of this composite score across all scenes.}
    
    \item \textcolor{r4color}{\textbf{Generality:} Although the hyper-parameters were not tuned on the SceneEval scenes, they perform effectively on the SceneEval benchmark, demonstrating their strong generalizability. Furthermore, the hyper-parameters only need to fall within a reasonable range to ensure convergence; the ultimate quality of the layout relies more heavily on the reasoning capabilities of the VLM and the core design of the hierarchical algorithm.}
    
    \item \textcolor{r4color}{\textbf{Stability Across Random Seeds:} We utilize the \textit{Optuna} framework\cite{akiba2019optuna} with a Tree-structured Parzen Estimator (TPE\cite{bergstra2011algorithms}) sampler to conduct the search. When repeating the search process with multiple different random seeds, although the internal exploration paths may vary, the final optimal parameters consistently converge to a similar, stable range, proving the robustness of our searched weights.}
\end{itemize}

\textcolor{r4color}{To enhance search efficiency, we employ the Hyperband pruner\cite{li2018hyperband} to terminate unpromising trials based on intermediate performance across scenes. The tuning process runs for 32 hours. The final optimized parameters used in our experiments are listed in \cref{tab:hyperparams}.}

\subsection{Object Retrieval Implementation Details}
\label{sec:retrieval_details}
\paragraph{3D Asset Database.}
Our system utilizes the 3D asset library created by Holodeck~\cite{yang2024holodeck}, which consists of over 51,000 diverse high-quality assets sourced from Objaverse~\cite{deitke2023objaverse}. These assets are specifically curated for indoor scenes and augmented with detailed metadata—including textual descriptions, bounding box dimensions, and canonical views—generated by GPT-4-Vision. Furthermore, the assets undergo optimization processes such as mesh reduction and collider generation to ensure efficient rendering and interaction within embodied AI simulators.

\paragraph{Retrieval Algorithm.}
To retrieve the most suitable 3D asset $O_i$ from the database given a textual query $T_q$ and a target geometric bounding box $B_{target} \in \mathbb{R}^3$, we employ a multi-stage scoring mechanism. The final relevance score $Score_{Final}(i)$ is a weighted sum of semantic, visual, and geometric similarities:

\begin{equation}
\begin{split}
Score_{Final}(i) & = w_{1} \cdot S_{sbert}(i) + w_{2} \cdot S_{clip}(i) \\
                 & \quad + w_{3} \cdot S_{size}(i)
\end{split}
\end{equation}

\noindent where $S_{sbert}(i)$ denotes the semantic similarity derived from SBERT embeddings (converted from $L_2$ distance in a coarse retrieval stage), and $S_{clip}(i)$ represents the visual similarity calculated via OpenCLIP. Specifically, $S_{clip}(i) = \max_{v \in \mathcal{V}_i} (\text{sim}(f_{txt}(T_q), f_{img}(v)))$, where $\mathcal{V}_i$ is the set of multi-view renderings of object $i$, taking the maximum similarity to handle viewpoint variations.

\paragraph{Geometric Alignment ($S_{size}$).}
To ensure the retrieved asset fits the planned layout not just in scale but in aspect ratio, we introduce a shape-aware geometric score. Direct comparison of dimensions is sensitive to absolute scale; therefore, we normalize the bounding box dimensions to focus on the object's proportions.

Let the target dimensions be $R_{target} = (w_t, d_t, h_t)$ and the candidate asset dimensions be $R_{asset}^{(i)} = (w_a, d_a, h_a)$. We first normalize these vectors by their respective maximum dimension to obtain scale-invariant representations $\hat{R}$:

\begin{equation}
    \hat{R}_{target} = \frac{R_{target}}{\|R_{target}\|_\infty}, \quad \hat{R}_{asset}^{(i)} = \frac{R_{asset}^{(i)}}{\|R_{asset}^{(i)}\|_\infty}
\end{equation}

\noindent where $\|\cdot\|_\infty$ denotes the infinity norm (maximum component). We then compute the geometric discrepancy $D_{size}$ as the mean absolute difference between the normalized vectors:

\begin{equation}
    D_{size}(i) = \frac{1}{3} \sum_{k \in \{w,d,h\}} \left| \hat{R}_{target}^{(k)} - \hat{R}_{asset}^{(i,k)} \right|
\end{equation}

\noindent Finally, this discrepancy is converted into a similarity score $S_{size} \in (0, 1]$ using an exponential decay function with a sensitivity hyperparameter $k$:

\begin{equation}
    S_{size}(i) = \exp(-k \cdot D_{size}(i))
\end{equation}

\noindent This formulation ensures that objects with similar aspect ratios to the planned placeholder receive significantly higher scores, promoting physical plausibility in the generated scene.

\paragraph{Parameter Settings.}
Based on our empirical validation, we set the retrieval weights to $w_1 = 5$, $w_2 = 85$, and $w_3 = 10$ in the final re-ranking stage, as the coarse stage already filters semantically irrelevant candidates. The coarse retrieval stage recalls the top-$K=60$ candidates based on SBERT similarity. For the geometric alignment, the sensitivity factor is set to $k=10$.

\begin{figure}[thbp]
    \centering
    \includegraphics[width=0.96\linewidth]{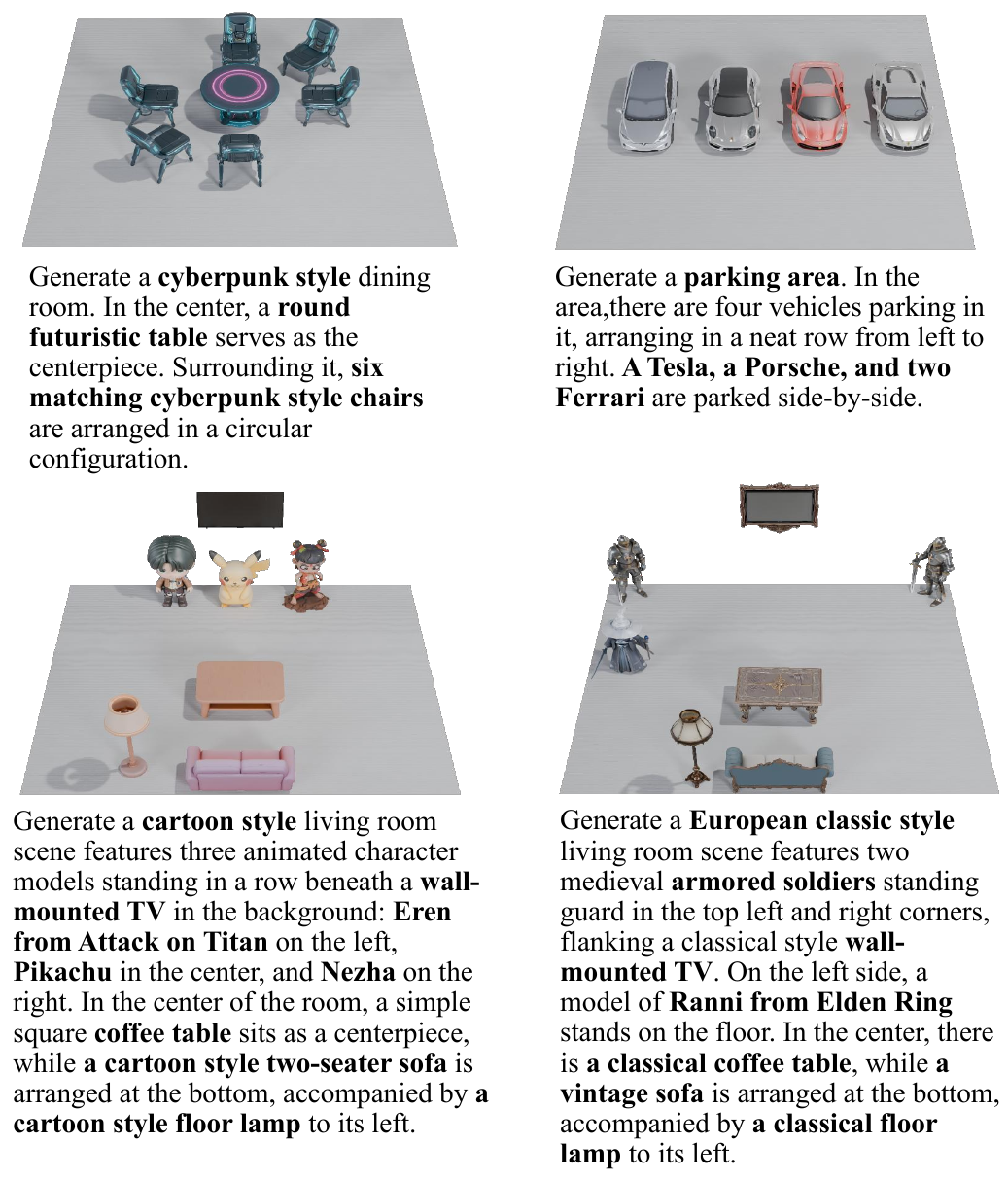}
    \caption{Examples generated by the pipeline with generative 3D objects. The pipeline supports generating assets that do not exist in the asset library, such as anime characters, sports car models, and more.
    }
\end{figure}

\subsection{Generation Pipeline with generative objects}

The modular design of our framework allows for the replacement of the retrieval-based asset acquisition with a generative approach. As illustrated in \cref{fig:pipeline_comparison}, instead of selecting an existing asset from a pre-defined database, we can synthesize a novel 3D object on the fly using a two-stage generative pipeline.

For example, in this alternative pipeline, the system first employs a Text-to-Image model (e.g., DALL-E) to generate a high-quality 2D image based on the textual description provided in the user query. Subsequently, this generated image serves as the input for an Image-to-3D model (e.g., Hunyuan 3D), which reconstructs the final 3D mesh geometry and texture. This generative method effectively addresses the limitations of fixed asset libraries, enabling the creation of highly specific or imaginative objects that may not be present in the retrieval database.

While the generative pipeline can produce more realistic and novel objects, it comes with a significant trade-off in processing speed compared to retrieval. Generating a usable 3D model typically requires at least 10 seconds, and for higher quality outputs, this can extend beyond two minutes. In contrast, retrieving an object from our optimized database takes less than one second. Given the emphasis on real-time scene editing and interactive performance in our system, the retrieval pipeline was chosen. However, for applications with less stringent real-time requirements, generative workflows offer a highly attractive alternative for scene creation and editing. In addition to the workflows we have demonstrated, 3D objects can also be generated directly from text information.

\subsection{Details of the RAG Template Library}

\textcolor{r4color}{The RAG template library contains common design rules, e.g., "large spaces should adopt a zoning-based layout to enhance spatial hierarchy." These rules were distilled from literature and public interior design resources (not from any datasets) and summarized as a template. The template is then chunked and embedded into a vector database using Qwen3-Embedding-4B. During inference, the system retrieves the top-3 relevant chunks per user query and incorporates them into the prompt to guide scene planning.}

Regarding the trade-off between diversity and plausibility, the RAG module provides general design rules rather than fixed, rigid templates, ensuring that layouts follow general design principles rather than merely memorizing dataset samples. Consequently, this approach minimally impacts generative diversity. Furthermore, we have successfully utilized extensive open-vocabulary instructions in our experiments, which demonstrates that our method effectively preserves both layout plausibility and scene diversity.

\subsection{Prompt for the Scene Planning Module}
The system prompt for the scene planning module is as follows; we use it to call the large model to generate the macro-planning of the scene and the list of required objects:

\styledfileinput[]{program/planning.txt}{}

After the large model outputs JSON-formatted text representing planning information, our system parses the model's output. For the asset list field, we split each object with the same name. For every ``\texttt{asset\_name}'', we generate distinct ``\texttt{object\_ids}'' based on its quantity. For example, if the ``\texttt{quantity}'' of ``\texttt{asset\_name}'' ``\texttt{book}'' is 4, the system automatically creates four ``\texttt{book}'' objects with ``\texttt{object\_ids}'' ``\texttt{book-0}'', ``\texttt{book-1}'', ``\texttt{book-2}'', and ``\texttt{book-3}''. These four objects share the same metadata but have different object IDs.

The user prompt for the scene-planning module is as follows:

\styledfileinput[]{program/planning_user.txt}{}

\subsection{Prompt for the Layout Generation Module}

The system prompt for the layout generation module is as follows; we use it to invoke the LLM to generate object placement coordinates and spatial relationships for the scene:

\styledfileinput[]{program/prediction.txt}{}

\section{Prompt for the Evaluation of SP Score}

In the experiment, we used GPT-5 to evaluate the semantic plausibility score metric. We input the top-down view of each scene, employed GPT-5, and combine it with generation instructions to score the scene’s top-down view on a 0–100 scale. The prompt used during the evaluation was as follows: 

\styledfileinput[]{program/evaluation.txt}{}

\begin{table}[tb]
\centering
\footnotesize
\caption{Quantitative results of scene editing. We report the Editing Success Rate (ESR), Scene-level Collision ($COL_{scene}$), and Scene-level Out-of-Bounds ($OOB_{scene}$).}
\label{tab::editing}
\begin{tabular*}{\linewidth}{@{\extracolsep{\fill}}l c c c} 
\toprule
\toprule
Instruction Type & ESR ($\uparrow$) & COL$_{scene}$ ($\downarrow$) & OOB$_{scene}$ ($\downarrow$) \\ 
\midrule
Move             & 80\%    & 10\%           & 10\% \\
Delete           & 100\%   & 0\%            & 0\% \\
Add              & 100\%   & 0\%            & 0\% \\
\bottomrule
\bottomrule
\end{tabular*}
\end{table}

\section{More results}
\subsection{Quantitative Results of Scene Editing}
\label{sec:editing_quant}

To comprehensively evaluate the performance of our editing module, we conduct a quantitative analysis focusing on execution accuracy and physical plausibility.

We employ an automated rule-based script to calculate the \textbf{Success Rate (ESR)}, ensuring objective verification without manual intervention. To evaluate spatially ambiguous instructions (e.g., ``add to the table'' or ``move to the corner''), our scripts check \textbf{geometric constraint satisfaction} rather than exact coordinates. Specifically, for \texttt{Add} and \texttt{Move} operations, the system automatically verifies if the target object establishes the correct spatial relationship (e.g., bounding box overlap or vertical support) with the reference region specified in the text.

Furthermore, we report \textbf{Scene-level Collision ($COL_{scene}$)} and \textbf{Scene-level Out-of-Bounds ($OOB_{scene}$)} to measure physical plausibility. Since each editing instruction corresponds to a single scene execution, we calculate these metrics as the percentage of \textit{edited scenes} containing at least one violation. This strictly reflects whether the instruction yields a usable layout. It is worth noting that all initial scenes are guaranteed to be collision-free and within bounds, ensuring these metrics strictly measure errors introduced during the editing process.

\cref{tab::editing} presents the quantitative results of our editing experiments. Our method achieves a 100\% success rate for \texttt{Delete} and \texttt{Add} instructions with zero introduced collisions or boundary violations. 

For \texttt{Move} instructions, the method maintains a high success rate of 80\% with low violation rates (10\% for both collisions and OOB). The slight performance drop compared to other operations is primarily attributed to the inherent ambiguity of natural language spatial descriptions. Instructions such as ``move the chair near the table'' lack precise metric definitions, occasionally leading to generated positions that are semantically reasonable but geometrically borderline under our strict verification protocols. Additionally, repositioning objects within cluttered layouts imposes tighter physical constraints than deletion or simple addition, increasing the difficulty of finding collision-free solutions.

\subsection{Qualitative Results}
\begin{figure*}[t]
    \centering
    \includegraphics[width=1\textwidth]{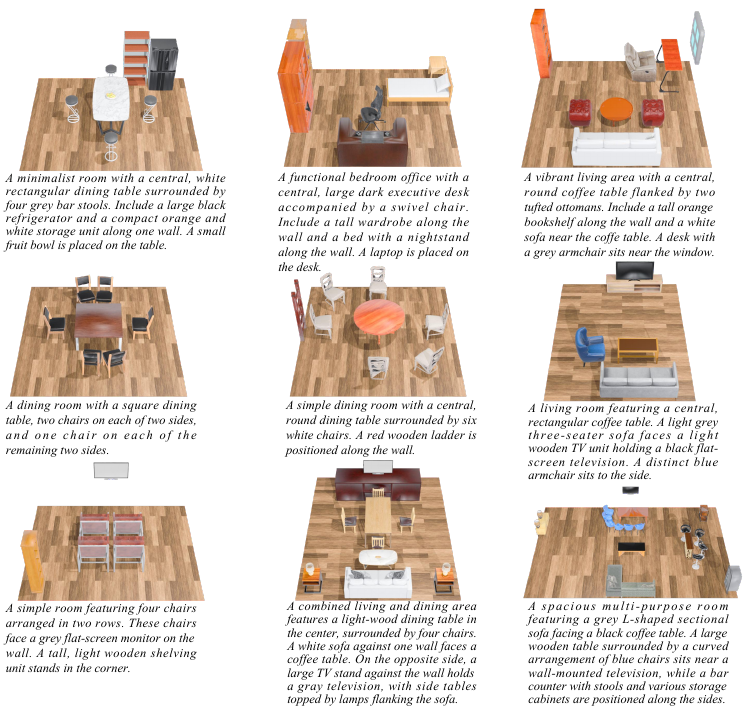}
    \caption{Layouts generated by HOG-Layout and the corresponding input instructions. 
    }
    \label{fig:appendix-qualitative}
\end{figure*}

\begin{figure*}[thbp]
    \centering
    \includegraphics[width=1\textwidth]{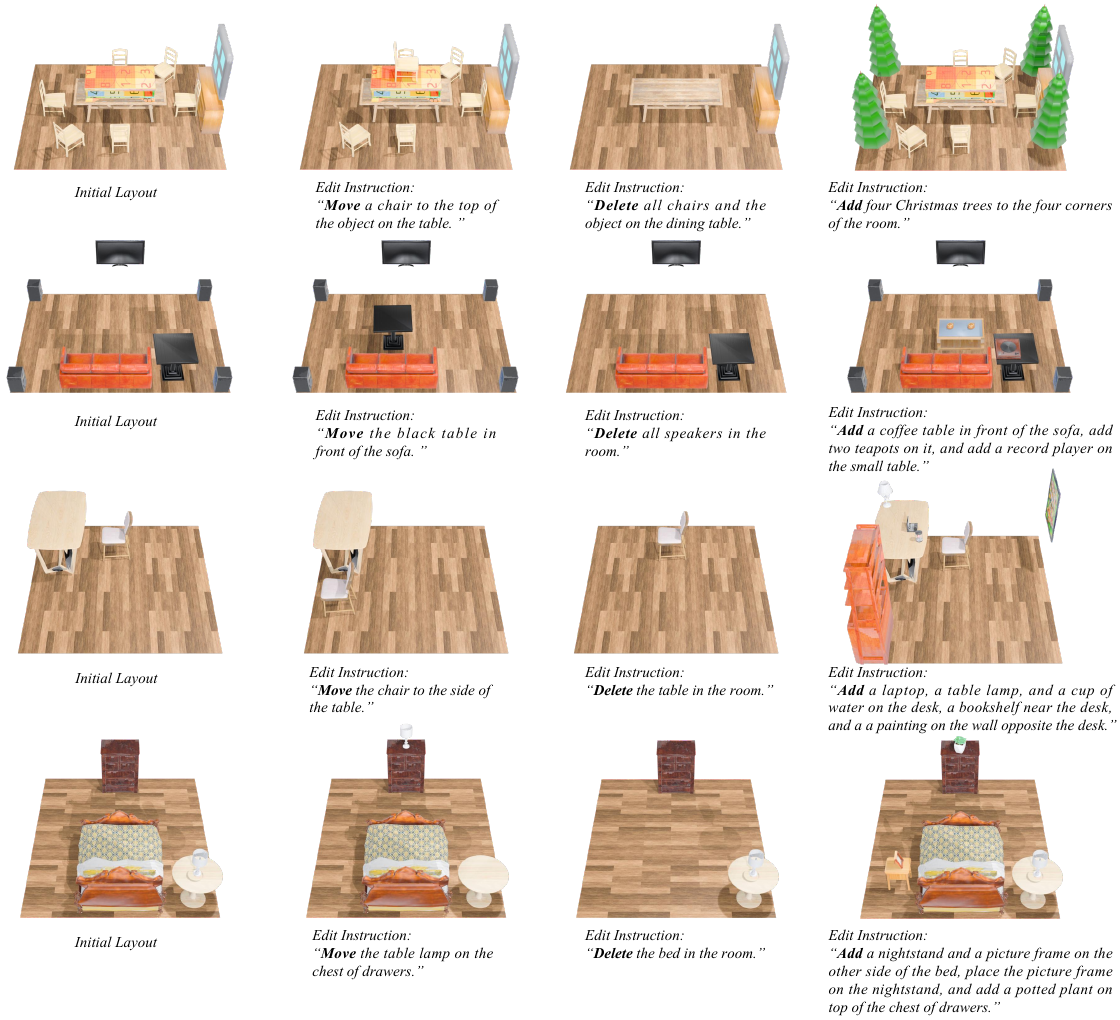}
    \caption{Edited layouts produced by HOG-Layout using the instructions.}
    \label{fig:appendix-editing}
\end{figure*}

\begin{figure*}[thbp]
    \centering
    \includegraphics[width=1\textwidth]{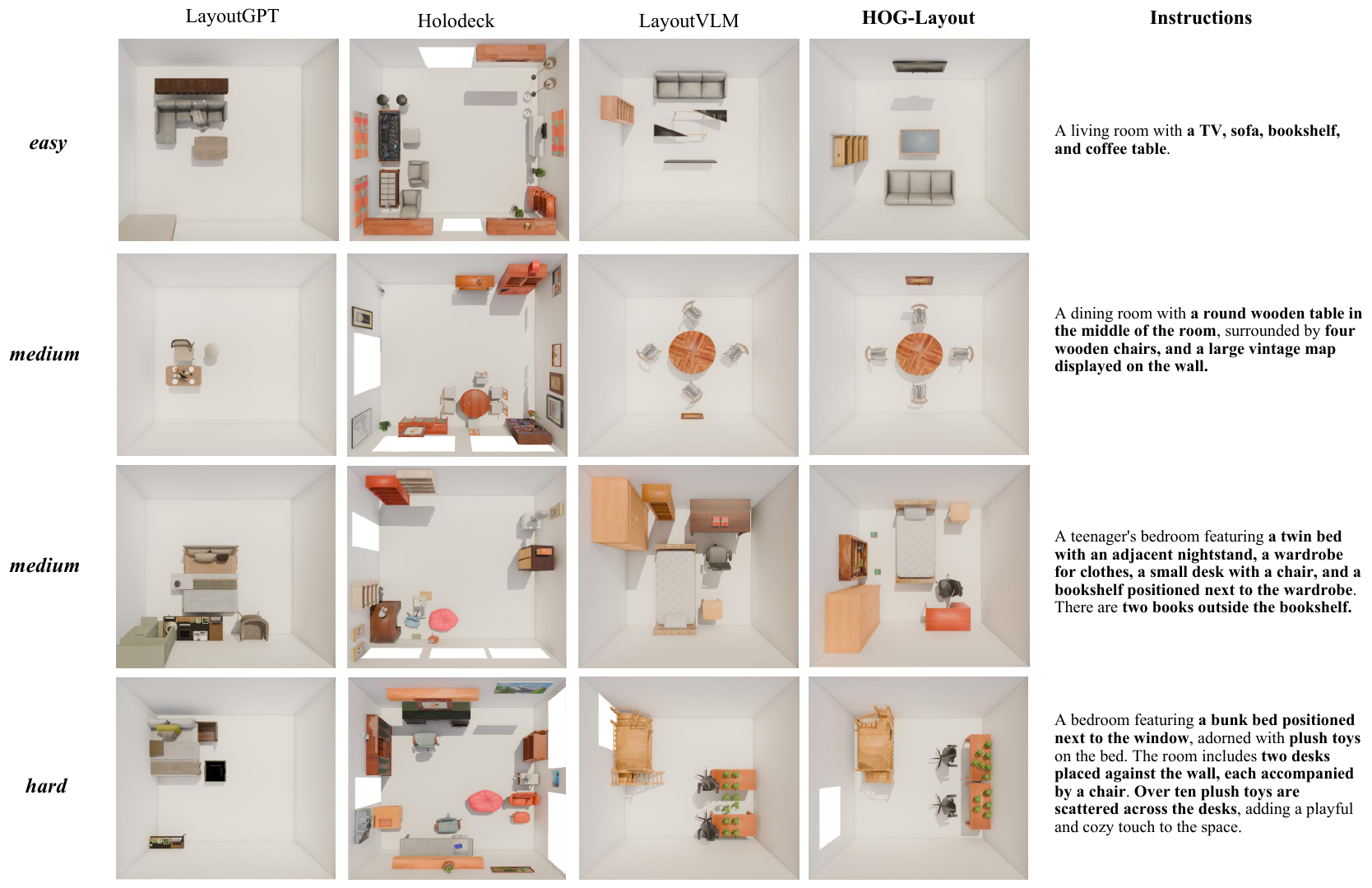}
    \caption{Generation examples of different methods on the benchmark.}
    \label{fig:compareg}
\end{figure*}

\cref{fig:appendix-qualitative}, \cref{fig:appendix-editing} and \cref{fig:compareg} show more qualitative results. \cref{fig:appendix-qualitative} and \cref{fig:compareg} are some generation results, while \cref{fig:appendix-editing} are editing results.

\end{document}